# HippMetric: A skeletal-representation-based framework for cross-sectional and longitudinal hippocampal substructural morphometry

Na Gao[1], Chenfei Ye[2], Yanwu Yang[3,4], Anqi Li[1], Zhengbo He[1], Li Liang[1], Zhiyuan Liu[5], Xingyu Hao[6], Ting Ma[6,7,8*], Tengfei Guo[1*]

[1]Institute of Neurological and Psychiatric Disorders, Shenzhen Bay Laboratory, Shenzhen, China
[2]International Research Institute for Artificial Intelligence, Harbin Institute of Technology (Shenzhen), Shenzhen, China.
[3]Department of Psychiatry and Psychotherapy, University Hospital Tübingen, Tübingen, Germany
[4]German Center for Mental Health (DZPG) , Germany
[5]Department of Computer Science, University of North Carolina at Chapel Hill, US
[6]School of Electronic & Information Engineering, Harbin Institute of Technology (Shenzhen), Shenzhen, China.
[7]Peng Cheng Laboratory, Shenzhen, China.
[8]Guangdong Provincial Key Laboratory of Aerospace Communication and Networking Technology, Harbin Institute of Technology (Shenzhen), Shenzhen, Guangdong, China
*Corresponding author:
Ting Ma. Phone: +86-755-26033608; e-mail: tma@hit.edu.cn
Tengfei Guo. Phone: +86-0755-2684-9264; e-mail: tengfei.guo@szbl.ac.cn

## Abstract

Accurate characterization of hippocampal substructure is crucial for detecting subtle structural changes and identifying early neurodegenerative biomarkers. However, high inter-subject variability and complex folding pattern of human hippocampus hinder consistent cross-subject and longitudinal analysis. Most existing approaches rely on subject-specific modelling and lack a stable intrinsic coordinate system to accommodate anatomical variability, which limits their ability to establish reliable inter- and intra-individual correspondence. To address this, we propose HippMetric, a skeletal representation (s-rep)-based framework for hippocampal substructural morphometry and point-wise correspondence across individuals and scans. HippMetric builds on the Axis-Referenced Morphometric Model (ARMM) and employs a deformable skeletal coordinate system aligned with hippocampal anatomy and function, providing a biologically grounded reference for correspondence. Our framework comprises two core modules: (1) a skeletal-based coordinate system that respects the hippocampus' conserved longitudinal lamellar architecture, in which functional units (lamellae) are stacked perpendicular to the long-axis, enabling anatomically consistent localization across subjects and time; and (2) individualized s-reps generated through surface reconstruction, deformation, and geometrically constrained spoke refinement, enforcing boundary adherence, orthogonality and non-intersection to produce mathematically valid skeletal geometry. Extensive experiments on two international cohorts demonstrate that HippMetric achieves higher accuracy, reliability, and correspondence stability compared to existing shape models. For example, HippMetric yields substantially lower cross-sectional and longitudinal correspondence errors than cm-rep, ds-rep, and SPHARM-PDM (≈1.6-1.7 mm for HippMetric versus 3.5-10.6 mm for alternative models), enabling more precise morphological analysis. Furthermore, HippMetric revealed Alzheimer's disease-related substructural alterations and improved prediction performance over conventional volumetric measures, highlighting its potential for robust, large-scale biomarker discovery.

## Graphic Abstract

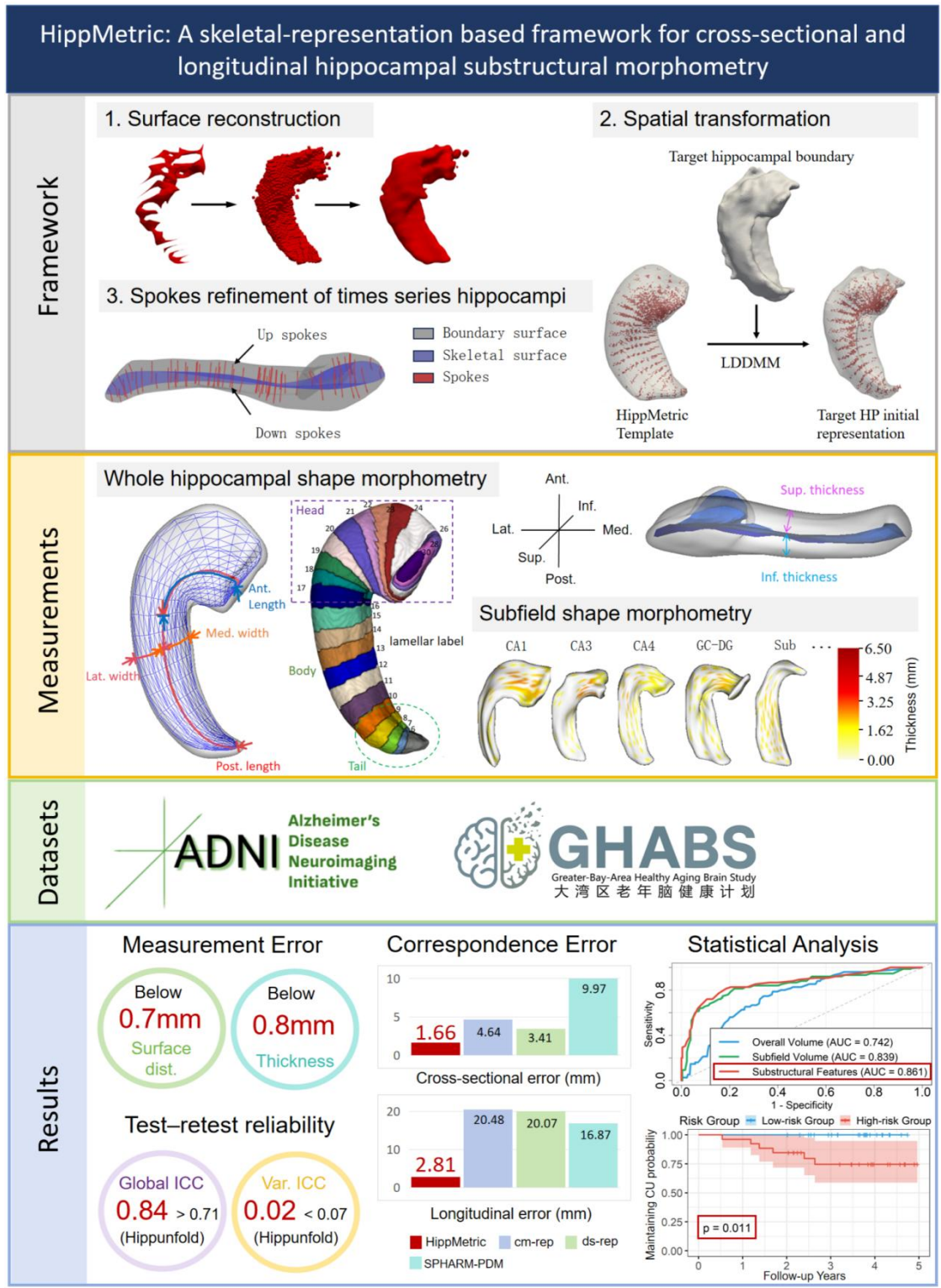

## Key words

## 1. Introduction

The hippocampus of human brain plays a critical role in memory, cognition, and neurodegenerative processes, and has long been a key target in imaging-based studies of structural alterations and early disease markers (Shahid SS et al., 2022; Vockert N et al., 2024; Haast RAM et al., 2024; Yushkevich PA et al., 2024). Traditional morphometric analyses have typically treated the hippocampus as a unified structure or divided it into discrete subfields based on histological atlases. While such approaches have advanced our understanding of gross anatomical changes, they fail to capture the local thickness of the folded archicortical layers and morphometry along the hippocampal long-axis, both of which are increasingly recognized as critical for functional specialization and differential vulnerability to disease (Strange BA et al., 2014; Stark SM et al., 2021; Ayhan F et al., 2021; Levone BR et al., 2021; Nichols ES et al., 2023; Vogel JW et al., 2020; Xie H et al., 2024).

Emerging morphometric tools have begun to incorporate the intrinsic geometry of the hippocampus into morphological analyses. Classical shape modeling methods, such as SPHARM-PDM (Paniagua B et al., 2012), cm-rep (Hong S et al., 2018) and s-rep (Pizer SM et al., 2022), modeling hippocampal structures by spherical harmonics, medial or skeletal representations (s-reps), offering compact shape descriptors and smooth correspondences across subjects. While those models showed promise in statistical shape analysis, they overlook the hippocampus's intrinsic anatomical and functional organization, such as the subfield-specific cytoarchitecture, and functional arrangement along the long-axis that governs information flow (Genon S et al., 2021), so that the geometric correspondences they establish are not grounded in the underlying anatomical organization. Therefore, shape differences captured by these models may not map onto biologically meaningful substructures or known disease-vulnerable pathways, limiting their interpretability and utility in mechanistic studies (e.g., subfield-specific vulnerability or resistance to pathology). This limitation is particularly challenging in neurodegenerative diseases, which are characterized by subfield-specific atrophy patterns grounded in histology (e.g., CA1 and subiculum in Alzheimer's disease (AD) (Braak and Braak, 1997; Kerchner et al., 2010)), and by a characteristic distribution of degeneration along the long axis, with the posterior often affected earlier (Adler, D.H et al., 2018). Capturing such point-wise, anatomically meaningful trajectories of degeneration remains largely beyond the reach of existing geometric models.

More recently, anatomically motivated frameworks have sought to better capture the internal organization of the hippocampus. For instance, Hippunfold establishes a two-dimensional coordinate system by flattening the hippocampal surface, with recent versions automating the identification of key subfield boundaries (DeKraker J et al., 2018; DeKraker J et al., 2022). Another approach, Hipsta constructs a three-dimensional coordinate system from tetrahedral volumetric meshes to estimate local thickness and curvature (Diers K et al., 2023). These methods improve spatial interpretability by providing local morphometric measures, in contrast to global volumetric metrics. However, a fundamental constraint persists: their coordinate systems are inherently boundary-driven. Whether boundaries are defined manually or identified automatically via image intensity, the entire geometric framework is anchored to the often ambiguous and variable tissue interfaces. As a result, the derived representations and morphometric estimates remain sensitive to segmentation instability, scan-rescan noise, and algorithmic choices in preprocessing. This inextricable link to boundary definition propagates uncertainty into quantitative measurements, undermining reproducibility and confounding the detection of subtle longitudinal change.

Furthermore, when applied to longitudinal analysis, the boundary-dependence of the current methods poses a specific conceptual challenge: to establish correspondence across time, these methods typically enforce a topology-preserving mapping between surfaces from different scans. This approach implicitly assumes the underlying tissue is continuously deformed, thereby mathematically maintaining a one-to-one correspondence for every observations, including regions that may have substantially atrophied or biologically disappeared in neurodegenerative progression. Consequently, they are ill-suited to model or measure true tissue loss, as their framework cannot naturally represent the diminishing or vanishing of structural elements. All above limitations highlight the critical need for a paradigm that moves beyond boundary anchoring, instead deriving its coordinate system directly from the hippocampus's intrinsic, internal architecture to ensure robustness and anatomical consistency.

Multimodal evidence from structural and functional imaging, transcriptomics, and cytoarchitecture reveals that the hippocampus exhibits a lamellar, sheet-like internal structure aligned with its crescentic, head-to-tail long-axis (Andersen P et al., 1971; Andersen P et al., 2000; Pak S et al., 2022; Sloviter RS and Lomo T, 2012; McCrea M et al., 2025). This anatomical organization supports topographically organized projections and parallel information processing pathways. This observed consistency in lamellar topology across individuals, despite variations in size, curvature, or digitations, suggests that a coordinate system grounded in these intrinsic anatomical features, rather than in variable boundary surfaces, could provide a reliable basis for establishing cross-subject correspondence. This lamellar organization can be captured by the framework of s-reps, where a field of radial spokes projects from the skeletal surface toward the boundary. A key advantage of this representation is that by defining shape from the interior scaffold outward, such models inherently distinguish the stable "skeleton" from boundary-related noise, providing theoretical robustness against segmentation variability (Pizer SM et al., 2012; Tu L et al., 2016, 2018). Furthermore, the s-rep shifts the local coordinate system from the complex boundary surfaces to the simpler and more stable internal skeleton. This shift enables direct and robust indexing of local morphological changes without requiring global reparameterization, which offers a particular strength for longitudinal quantification.

Building on those insights, we previously introduced the Axis-Referenced Morphometric Model (ARMM) (Gao N et al., 2024). ARMM implements the s-rep through a coordinate system built upon the hippocampal skeletal surface, such that any point within the hippocampus can be indexed by its position on this surface and the associated spoke. By anchoring anatomical correspondence within its interior framework, ARMM reduces sensitivity to unreliable boundary definitions arising from segmentation inconsistencies or noise. While ARMM established a theoretically robust foundation, its full potential was constrained by two practical limitations: it did not support detailed subfield-level morphometry, and its correspondence performance had not been directly validated.

Building on the ARMM framework, we introduce HippMetric, a computational framework for precise and comprehensive hippocampal morphometry. HippMetric retains ARMM's interior skeletal coordinate system but extends it in several key directions to enable anatomically grounded subfield-level and longitudinal analyses within a common space. Specifically, we incorporate: (1) a subfield-aware thickness mapping strategy that integrates histological priors to maintain lamellar consistency without relying on surface reparameterization; (2) a cross-sectional and longitudinal alignment procedure that operates directly within the skeletal coordinate system, allowing anatomically matched comparison of local morphology across individuals and time; and (3) a geometry-preserving regularization mechanism that maintains skeletal properties during modelling, ensuring that derived measures such as thickness and width are geometrically rigorous and anatomically interpretable. These components form an integrated framework that supports estimation of hippocampal substructural features. We validated HippMetric through a series of quantitative benchmarks, including agreement of thickness estimates with manual annotations, reproducibility assessments, robustness evaluations in disease, etc. The validation experiments were conducted using subjects drawn from the North American aging population-based ADNI cohort and the South China aging population-based GHABS cohort, ensuring that performance was assessed across distinct populations and imaging conditions. Together, these advances constitute a methodological framework that provides a coherent set of quantitative metrics for hippocampal substructure, enabling sensitive detection of subtle and distributed morphological alterations in both cross-sectional and longitudinal neuroimaging studies. In summary, our contributions are as follows:

1. We propose HippMetric, an automated and anatomically grounded framework that jointly models the hippocampus across its subfield architecture and long-axis, enabling both cross-sectional and longitudinal morphometry within a unified representation.

2. HippMetric leverages a skeletal-based representation with geometry-constrained refinement that enforces boundary adherence and valid spoke geometry, enabling accurate and highly reproducible subfield thickness estimation that surpasses boundary-driven approaches.

3. HippMetric supports robust inter-subject and longitudinal alignment by embedding all scans into an anatomy-motivated intrinsic skeletal coordinate system, yielding stable topological correspondence over time. These substructural features consistently outperform conventional volumetric metrics in distinguishing AD and predicting clinical progression across two international cohorts.

## 2. Related Works

This section reviews hippocampal morphometry methods spanning from global volumetric analyses to surface-based and subfield-level approaches. We highlight recent advances that leverage intrinsic geometric modeling to enhance anatomical accuracy and consistency of alignment across individuals. Finally, we discuss the limitations of current methods and emphasize the need for multi-scale, anatomically informed frameworks capable of reliably detecting subtle, localized morphological changes.

**2.1 Surface-based morphometry and its hippocampal alignment framework**

Surface-based morphometry is widely used in hippocampal shape analysis, aiming to detect localized morphological variation through geometric alignment across individuals. A key challenge is to establish accurate and stable point-wise correspondence on anatomically variable hippocampal surfaces.

Registration-based approaches, such as Large Deformation Diffeomorphic Metric Mapping

(LDDMM) (Huang et al., 2006; Miller et al., 2002) and conformal mapping (Shi et al., 2013), align hippocampal surfaces via global deformation or intrinsic curvature-driven mappings. Elastic shape analysis frameworks, including square-root normal fields (SRNFs) and Q-maps (Hamid et al., 2017), define Riemannian metrics on shape space and optimize for minimal elastic deformation. Other intrinsic descriptors, such as the heat kernel signature (HKS) (Wangi et al., 2015; Feng et al., 2024) capture fine-scale local geometry and support isometry-invariant matching. Point distribution models (PDMs) and their spherical variant (SPHARM-PDM) (Styner et al., 2006; Fang et al., 2019) parameterize surfaces using point sets or spherical harmonics, enabling compact representations of shape variation.

However, surface-based methods are defined entirely by boundary geometry, making them highly sensitive to small segmentation imperfections—an issue that is particularly pronounced for the hippocampus, where limited image resolution often introduces boundary noise. As a result, these approaches can produce unstable or biologically implausible correspondences, especially in regions with complex folding or across longitudinal scans. This limitation motivates anatomically informed models that incorporate interior coordinate systems aligned with hippocampal organization.

**2.2 Skeletal-based shape modeling**

Skeletal–based representations offer a robust alternative to surface-based methods by anchoring shape analysis to intrinsic, centrally located structures that capture global geometry and symmetry. Unlike surface models, which are highly sensitive to segmentation errors and the individually variable digitations at the hippocampal head and tail, skeletal approaches leverage the relatively consistent internal architecture of the hippocampus. In particular, the primary curvature along the longitudinal axis generally follows a characteristic crescent-shaped trajectory, providing a stable reference for point-wise correspondence and morphometric quantification. Over the past two decades, various skeletal modeling techniques have been developed, including ridge-following methods, boundary erosion, Voronoi-based skeletonization, and continuous medial representations (cm-reps) (Yushkevich et al., 2006; Pouch et al., 2012).

Among the most influential are s-reps (Damon, 2003; Pizer et al., 2003) and their statistical refinements (Tu L et al., 2016, 2018), which model shape using skeletal points and associated spokes. Recent developments, such as ds-reps (Liu et al., 2023), introduces automated skeletal fitting within diffeomorphic frameworks, enabling discrete skeletal modeling. Further developments within this skeletal modeling paradigm have focused on enhancing mathematical rigor, such as the Elliptical Tube Representation (ETRep) which enforces geometric validity constraints for population-level shape analysis (Taheri Shalmani et al., 2025). Such approaches focus on statistical shape descriptors regardless of the actual anatomical organization of the hippocampus.

To address this, we previously introduced the multi-scale skeletal representation (m-s-rep) (Gao et al., 2023; Gao et al., 2024a), a hippocampus-specific skeletal model that incorporates biologically meaningful internal axes reflecting its longitudinal architecture, supporting multiscale morphometry including global shape and point-wise quantification. Nonetheless, m-s-rep is restricted to genus-zero structures and generally applied to the whole hippocampal shape, limiting its capacity to capture subfield heterogeneity.

**2.3 Hippocampal subfield morphometry**

Recent advances also extend hippocampal morphometry from whole-structure to subfield-level analysis, leveraging intrinsic anatomical constraints and surface-based coordinate systems. These methods aim to improve spatial specificity and interpretability in detecting regionally localized changes.

Hippunfold introduces an anatomically grounded unfolding framework, establishing a boundary-based coordinate system (from anterior-posterior, proximal-distal and laminar directions) via Laplace's equation on high-resolution MRI (DeKraker et al., 2018; DeKraker et al., 2022). This unfolded space allows consistent 2D mapping of subfields and quantification of features such as thickness, curvatures and gyrification. Its automated software package integrates deep learning-based segmentation with topologically constrained surface reconstruction, supporting population-level analyses. Hipsta (Diers K et al., 2023) constructs a tetrahedral mesh representation of the hippocampus endowed with a three-dimensional intrinsic coordinate system (e.g., capturing medial-lateral, anterior-posterior, and depth dimensions). By establishing this continuous volumetric coordinate frame, rather than flattening the surface to a 2D plane, Hipsta directly computes morphometric properties such as thickness along anatomically coherent directions.

While Hippunfold and Hipsta differ in their technical implementations, by using 2D unfolding and 3D volumetric parameterization respectively, both construct their coordinate systems from the boundary inward. This boundary-centric paradigm inherently ties geometric correspondence to segmentation quality and boundary noise, motivating the need for a framework that derives its coordinates from the interior outward.

**2.4 The skeletal coordinate system of hippocampal lamellar architecture**

Building on the advantages of skeletal-based representations (Section 2.2) and anatomically informed subfield-level hippocampal modeling (Section 2.3), we previously proposed the Axis-Referenced Morphometric Model (ARMM) (Gao et al., 2024b). ARMM constructs a skeleton along the hippocampal long-axis and defines a diffeomorphic transformation from a population template to individuals, which established anatomical-informed point-wise correspondences. Within this intrinsic space, ARMM enables multiscale quantification of local thickness, width, curvature, digitations, and regional-level metrics. Compared with surface-based methods and m-s-rep, ARMM improves reproducibility, biological validity, and sensitivity to AD-related alterations.

ARMM is motivated by evidence that the hippocampus consists of thin, sheet-like lamellae orthogonal to the longitudinal axis (Andersen et al., 1969; Andersen et al., 2000; Sloviter & Lømo, 2012), forming continuous topological units despite inter-individual variation (Ding & Van Hoesen, 2015). Axonal projections (~1.4mm) are largely confined within lamellae, while longitudinal pathways mediate inter-lamellar communication (Pak et al., 2022, McCrea M et al., 2025). Preserving this lamellar structure is critical for anatomically faithful morphometry and may provide a consistent basis for establishing point-wise correspondence across individuals, despite the large inter-subject variability.

However, ARMM was originally designed for global, cross-sectional analysis and does not directly support subfield thickness estimation or longitudinal alignment, which are necessary for tracking subtle, spatially localized changes over time. Extending ARMM to subfield-level modeling with intrinsic longitudinal alignment is therefore urgently required.

# 3. Method

We present HippMetric, a skeletal-based framework for quantitative morphometry of hippocampal substructures. The methodology is organized as follows: Section 3.1 describes the s-rep and associated geometric variables used in this paper. Section 3.2 outlines the overall HippMetric framework. The core methodological innovations are then detailed across three subsequent sections: Section 3.3 describes how the s-rep is used to establish anatomically consistent cross-subject and longitudinal correspondence; Section 3.4 explains the geometry-preserving regularization that ensures the mathematical validity and robustness of the skeletal model; and finally, Section 3.5 defines the subfield-aware thickness estimation that leverages this skeletal coordinate system for fine-grained analysis.

## 3.1 Preliminary of skeletal representation (s-rep)

The HippMetric is built upon the s-rep framework, which mathematically derives from the medial axis geometry. For a 3D object, its medial axis, or medial locus, is defined as the set of centers of all maximal inscribed balls (MIBs) within the object (Siddiqi et al. 2008). This axis, together with the corresponding ball radii, captures the object's intrinsic symmetry and volumetric structure. Formally, consider an unbranching genus zero 3D object. Let its medial manifold be $(m, r) \in R^3 \times R^+$, where $m$ is a continuous smooth surface in $R^3$, and $r$ is a scalar field, also called radial vectors, on $m$. Since any point on the boundary has only one corresponding MIB, the boundary surface can be completely reconstructed by MIBs. Let each point on the boundary $S$ correspond to a vector $\boldsymbol{r}$. $\boldsymbol{r}$ starts from $m$ and ends at $S$, and is perpendicular to $S$. The vector field $\boldsymbol{r}$ has three properties. Consider any point $p$ on $m$, it has two vectors $r^+, r^- \in \boldsymbol{r}$. Let the normal vector of $m$ at $p$ be $n_p$, $r^+$ point from p to $S_p^+ \in S$, and $r^-$ point from p to $S_p^- \in S$. Let the normal vectors of S at $S_p^+$ and $S_p^-$ be $n_{S_p^+}$ and $n_{S_p^-}$. Then 1) $<n_p, r^+> = <r^-, -n_p>$ , 2) $\|r^+\| = \|r^-\|$, and 3) $n_{S_p^+} \perp r^+, n_{S_p^-} \perp r^-$. The edge curve of medial surface, also called the fold curve, have the third vectors $\boldsymbol{r^0}$. A $r^0$ points from a fold point $m_0$ to $S^0 \in S$. Under these conditions, the boundary surface of the object $S$ is formed from three open manifolds: $S^+ = m + r^+$, $S^- = m + r^-$, $S^0 = m + r^0$. In medial geometry, the vector field $\boldsymbol{r}$ is also a radial map that transport the medial geometry to the boundary surface geometry, where Damon derives specific formula in literatures (Damon J et al., 2003; Damon J et al., 2004). The radial map guarantees 1-1 onto correspondence between the medial surface $m$ and the object boundary surface $S$.

Unlike medial methods that derive skeletal structure from the boundary, a process sensitive to noise and prone to unstable branching, the s-rep is explicitly formulated as a generative model that proceeds from the skeleton to the boundary. According to Damon (Damon J et al., 2003; Damon J et al., 2004), this s-rep formally generalizes the medial representation through the relaxed Blum condition, which constrains the length and orientation of spokes (analogous to radial vectors in medial representation) via three specific conditions:

(1) Boundary convergence: The terminal points of spokes do not precisely converge to the boundary surface.

(2) Orthogonality: The spokes do not precisely required to be orthogonal to their local tangent planes of the boundary at implied boundary points.

(3) Non-intersection: To ensure smooth interior geometry, spokes must not intersect locally. This ensures that each implied boundary point corresponds to a unique spoke, thereby guaranteeing the uniqueness of thickness values across surfaces.

Here, the constraints on boundary convergence and orthogonality are relaxed, while the non-intersection condition is enforced, which ensures the geometric regularity of the implied vector field throughout the interior of the object.

**3.2 Overview of the HippMetric framework**

The HippMetric framework consists of two main components: (1) template construction and (2) individual-level processing. The template is manually assisted and tailored according to the user-specified hippocampal subfield segmentation protocol. Its construction involves building skeletal surfaces within each subfield and generating boundary-oriented spokes, which are distributed according to the axes defined by the ARMM. Detailed procedures for template construction are described in Section 3.3.1.

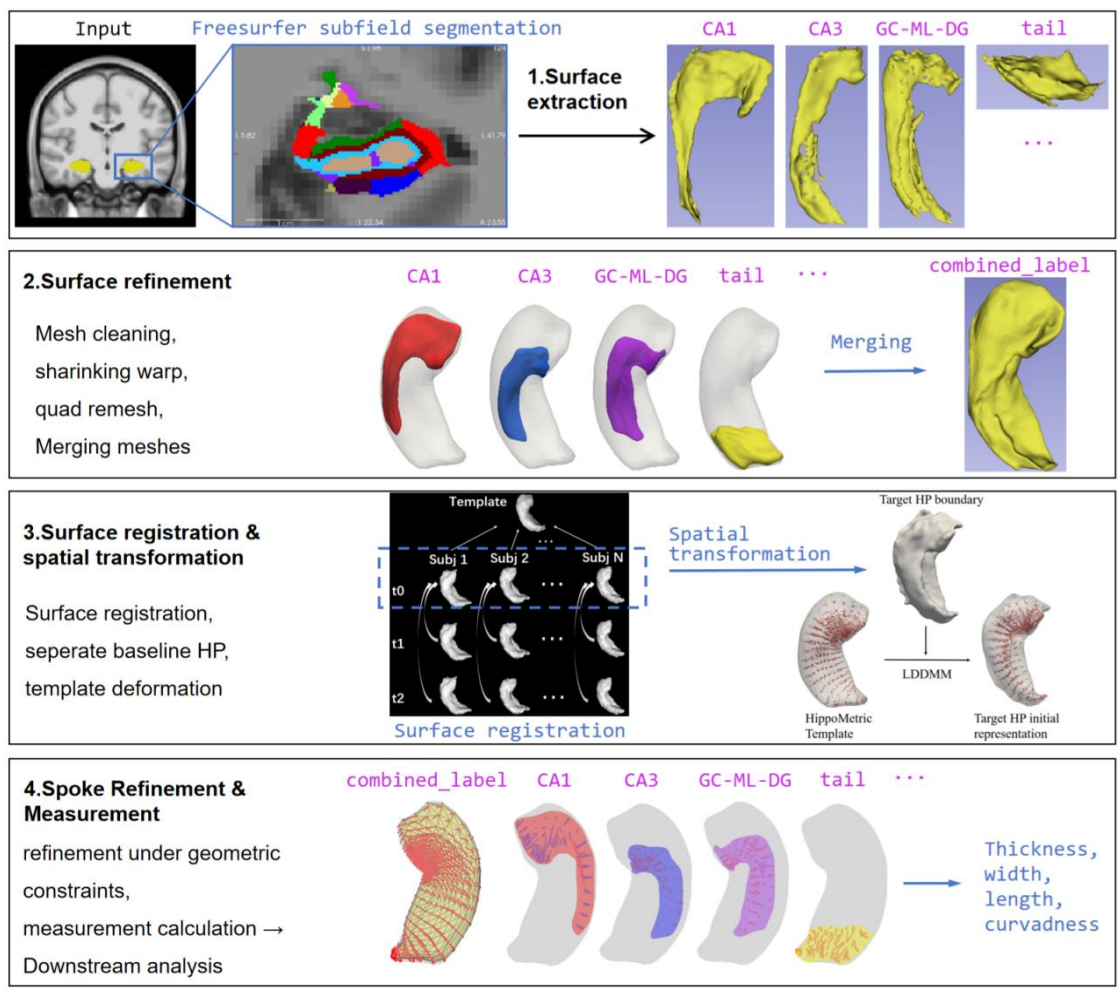

*Figure 1.* The HippMetric processing pipeline. The framework begins with (1) Surface Extraction, obtaining initial subfield meshes from FreeSurfer's FS60 hippocampal segmentation. (2) Surface Refinement follows, applying automated mesh cleaning, shrinking warp, quadrilateral remeshing, and merging to produce a single, topologically consistent hippocampal surface. For point-wise cross-sectional and longitudinal correspondence, (3) Surface Registration and Spatial Transformation aligns baseline surfaces to the template via LDDMM-based diffeomorphic registration, initializing a subject-specific s-rep. The follw-ups are then aligns to the baseline. Finally, individualized (4) Spoke Refinement optimizes spoke length and orientation under skeletal geometric constraints, from which morphometric features (thickness, width, length, curvedness, etc.) are computed.

Once the template is established, subject-level processing proceeds through four core stages (Figure 1). First, FreeSurfer's segmentation outputs are resampled to extract boundary surfaces, producing smooth and coherent surfaces for downstream processing. Second, surface refinement is performed using our automated pipeline, which addresses irregularities in the extracted surfaces, removes isolated meshes and holes, and merges individual subfields into a coherent hippocampal structure while preventing overlap. This stage includes steps such as mesh cleaning, shrinking warp and quad remeshing. Third, surface registration and spatial transformation are applied to establish cross-subject skeletal correspondence, which involves separation of baseline hippocampi, and template non-linear spacial transformation. Fourth, spoke refinement are performed under geometric constraints, ensuring skeletal integrity while deriving accurate morphometric measurements for downstream analyses.

Overall, this pipeline takes FreeSurfer's recon-all output as its input and generates individualized hippocampal morphological measures at global, subfield, and local levels. These measures are anchored in stable skeletal coordinate system with biologically informed correspondence, supporting robust and comprehensive cross-sectional and longitudinal analyses.

**3.3 Skeletal-based cross-sectional and longitudinal alignment**

The core of HippMetric's alignment strategy lies in its use of a unified skeletal coordinate system to establish correspondence across subjects and time. Cross-sectional alignment of interior geometry is achieved by diffeomorphically mapping individual hippocampi to a shared skeletal template, using a continuous, nonlinear deformation field. This field simultaneously warps the skeletal points and its associated boundaries from the template to the target. As a result, the arrangement of corresponding skeletal points is guaranteed to share a consistent spatial relationship to the hippocampus's internal axes of organization, defined by its long-axis and lamellar structure.

**3.3.1 Determination of template boundaries and s-rep**

The global coordinate system in HippMetric is inherited from the ARMM framework. It is established by constructing a continuous skeletal surface from the overall hippocampal shape, defining the long-axis on this surface, and reparameterizing it to create a lamellar distributed coordinate mesh. Spokes are then fitted at the mesh vertex and optimized under the partial Blum condition to form a geometrically consistent s-rep. We refer to the original ARMM work (Gao et al., 2024b) for full details of this process. HippMetric's contribution focuses on extending this stable global framework to enable detailed subfield-level analysis.

For each subfield surface, we constructed a Voronoi diagram by sampling boundary points, a method that is intrinsically linked to skeleton extraction. Only vertices fully enclosed within the surface were retained, forming a dense interior point cloud. We then fit this cloud with Non-Uniform Rational B-Splines (NURBS) to generate a smooth skeletal surface (Pizer SM et al., 2025) that approximates the subfield's internal geometry (Fig. 2). However, the CA1, GC-ML-DG, and molecular layer subfields, due to their complex geometry and folding structures, cannot be accurately represented by a single enclosed skeletal surface. These subfields were therefore each split into two anatomically defined parts, such as anterior/posterior or superior/inferior divisions, which we refer to as sub-subfields. Separate Voronoi sampling and NURBS fitting were applied to generate skeletal surfaces for each sub-subfield, which were subsequently merged into composite skeletal surfaces (Fig. 2).

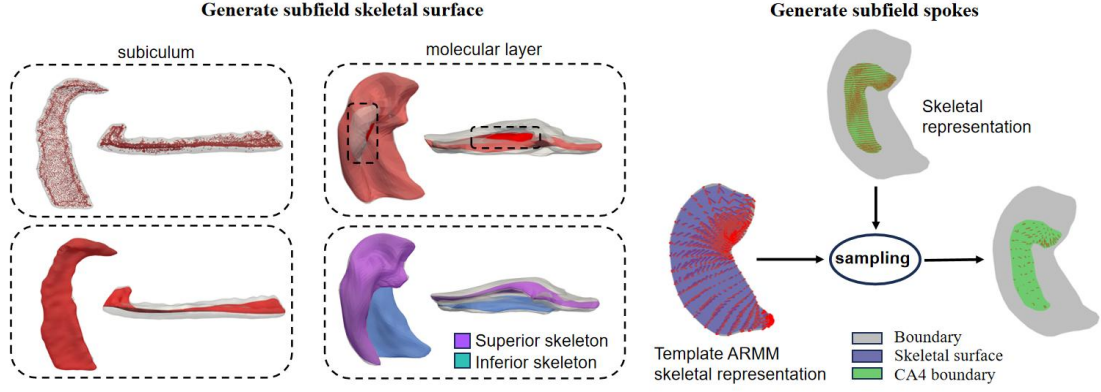

*Figure 2. Template construction based on the ARMM framework and the FS60 protocol. Left: Generation of the skeletal surface for each subfield, illustrated with a single skeletal surface in the subiculum and a composite skeletal surface in the molecular layer formed by merging two skeletal surfaces. Right: Lamellar-aligned sampling of subfield s-reps within the skeletal coordinate system.*

Based on each subfield's skeletal surface, we initialized a spoke field for thickness measurement. Each spoke originated from a vertex on the skeletal mesh, with its length and direction governed by the three principles outlined in Section 3.1. In practice, it is challenging to simultaneously enforce orthogonality and non-intersection. To resolve this, we introduce three penalty terms that jointly optimize spoke geometry while maintaining the core constraints. These terms minimize deviations from the ideal configuration.

Let $E_0$ quantify the Euclidean distance from each spoke $s$ to the closest vertex on the boundary surface. Specifically, for each spoke $s_i$, $E_0$ is defined as the minimum Euclidean distance between its terminal point $p_i$ and $v_j$:

$$E_0(s_i) = \min_{v_j \in S} \|p_i - v_j\|_2, \text{ where } v_j = arg \min_{v \in S} \|p_i - v_j\|_2,$$

where $S = \{v_1, v_2, ..., v_n\}$ is the set of surface vertices.

Let $E_1$ penalize the angular deviation between the spoke direction and the surface normal at the nearest boundary vertex, capturing the extent to which the orthogonality constraint is violated. Specifically, let $\mathbf{v}_i$ be the unit direction vector of spoke $s_i$, and $\mathbf{n}_i$ the unit normal vector at the closest boundary point. The angle between them is denoted by $\theta_i$, and $E_1$ is defined as:

$$E_1(s_i) = 1 - cos\,\theta_i = 1 - (\mathbf{v}_i \cdot \mathbf{n}_i)$$

Let $E_2$ measures the asymmetry in length between the superior and inferior spokes of each medial point. Let $s_i^{sup}$ and $s_i^{inf}$ denote the superior and inferior spokes, respectively. Then, $E_2$ is defined as the absolute difference of their lengths:

$$E_2(s_i) = \left| \|s_i^{sup}\|_2 - \|s_i^{inf}\|_2 \right|$$

To ensure geometric validity, the implied boundary points initially located outside the boundary surface are first projected inward onto the surface. Because our thickness measure is defined directly by the spoke length, the boundary convergence condition ($E_0$) must be strictly satisfied to guarantee measurement accuracy. Therefore, $E_0$ is enforced to align implied boundary points with corresponding boundary vertices. Subsequently, spoke directions are optimized to minimize angular deviation $E_1$. The sequence $E_0 \to E_1$ is executed iteratively for $n$ rounds (empirically, often $n = 2$ or 3 suffices). Then, the $E_2$ condition, which is derived from medial geometry, is applied after the last iteration to correct spokes with large angular deviations resulting from $E_1$ optimization, thereby ensuring local non-intersection and geometric regularity of the spoke field. As orientation updates may slightly displace endpoints, a final $E_0$ correction is applied to restore surface alignment. This process yields an initialized spoke field for each hippocampal subfield that meets the required geometric constraints.

Following spoke optimization, each subfield's s-rep was sampled according to the spokes defined by ARMM. Each ARMM skeleton point links to a superior and a inferior spokes. These spokes intersect the subfield skeletons, with intersection points matched to the nearest mesh vertices to define subfield-specific measurement locations. The right panel of Fig. 2 depicts the initialized CA4 spokes sampled at 100 skeletal vertices. Spoke endpoints (red arrow tips) closely conform to the gray boundary surface without self-intersections. Since each subfield vertex is uniquely indexed by an ARMM spoke, measurements are intrinsically aligned with the hippocampus's global lamellae, preserving inter-subject correspondence and enhancing comparability of subfield morphological features across individuals.

**3.3.2 Align template skeletal points to individuals**

To ensure spatial consistency, a rigid registration strategy aligns each subject's hippocampal surfaces to the template space. The registration is guided by the whole hippocampal surface, with the derived linear transformation applied uniformly to all subfield boundaries to maintain their spatial relationships. For longitudinal data, once the baseline scan is aligned to the template, all follow-up scans are rigidly registered to their corresponding baseline. The procedure maintains intra-subject consistency across time points while preserving alignment with the template.

To map the template s-rep model onto individual hippocampi, we employed a diffeomorphic surface registration strategy based on the Large Deformation Diffeomorphic Metric Mapping (LDDMM) framework. This method computes a smooth, invertible deformation field that transforms the template surface $S_{temp}$ to the subject-specific hippocampal surface $S_{subj}$, while preserving anatomical correspondences. Once the optimal transformation is estimated, it is applied to all internal template landmarks, including skeletal points and implied boundary points, to project them into subject space, preserving their spatial arrangement relative to the global surface. In practice, this procedure was implemented using *Deformetrica* in dense mode (Fishbaugh J et al., 2017). We iteratively refine the Gaussian kernel width in small steps to minimize surface mismatch, until the average surface distance fell below a predefined threshold.

For longitudinal analysis, we introduce a model that captures structural change whether it manifests as progressive erosion (tissue loss) or outward extension (tissue growth). This contrasts with surface- or boundary-based methods that establish correspondence by independently reparameterizing the entire surface mesh in each follow-up observation. Such independent reparameterization enforces a complete, point-wise mapping across time, which mathematically preserves a correspondence to regions even after they have atrophied or changed substantially, thereby failing to capture true tissue loss. To avoid this, our approach models follow-up scans within the same individualized skeletal coordinate system established at baseline, directly deforming its baseline skeletal representation. For atrophy, morphological change is represented as a controlled inward erosion of the skeletal surfaces coupled with spoke shortening. For growth, the skeletal surfaces remain stable while spoke lengths increase. This allows thickness measures to approach zero for severely atrophied regions or increase for expanded regions, all within a stable, individualized coordinate framework that ensures consistent and robust quantification of longitudinal change.

**3.4 Geometry-preserving regularization of skeletal geometry**

S-reps requires that the spokes satisfy three fundamental geometric constraints: boundary convergence, orthogonality, and non-intersection. Raw boundaries derived from voxel-level segmentations are often irregular, non-smooth, and may contain isolated components or holes, which can adversely affect the fidelity of shape modeling. To address this, the extracted surfaces are refined to produce coherent and topologically plausible boundaries. The lengths and directions of the spokes are then iteratively adjusted to satisfy the three constraints.

**3.4.1 Surface refinement**

Ten hippocampal subfields were selected from the Freesurfer segmentation pipeline with the FS60 protocol: CA1, CA3, CA4, GC-ML-DG, molecular layer, HATA, subiculum, parasubiculum, presubiculum, and tail. The Fig. 1B illustrates the hippocampus and subfield locations based on the FS60 atlas.

All segmentation results were resampled to 0.5 mm isotropic resolution, balancing surface smoothness and computational efficiency. Hippocampal subfield surfaces were reconstructed by extracting voxel-wise segmentation boundaries via the Visualization Toolkit (VTK). Subsequently, the ten subfield surfaces were concatenated to form a composite whole-hippocampus morphology, which differs from the whole hippocampus segmentation produced directly by FreeSurfer recon-all pipeline.

Raw surface meshes extracted from segmentation labels often suffer from topological inconsistencies, such as small disconnected components and holes (Fig. 3A), due to imperfections in voxel-wise labeling. To improve anatomical validity and geometric quality, a three-stage refinement pipeline was implemented. First, each mesh was decomposed into connected components, and only the largest anatomically plausible component was retained based on vertex count, effectively removing spurious fragments. Next, a shrink-wrapping strategy was applied to produce watertight surfaces while preserving local geometry. This step included automated hole filling and adaptive surface offsetting, with mesh resolution controlled by a target edge length proportional to subfield complexity, for example, using finer resolution for the dentate gyrus and coarser for the subiculum. Finally, the surfaces were uniformly remeshed using a quadrilateral-based scheme to enhance face regularity and ensure numerical stability in subsequent analyses. The number of target quads was empirically adjusted per subfield to balance anatomical detail and computational cost. All refined meshes were exported with triangulated faces to ensure compatibility with subsequent morphometric processing. To facilitate the subsequent description, we designate this procedure as the Robust Boundary Optimization Scheme (RBOS).

In total, ten subfields were processed independently, each undergoing separate surface refinement. As subfield boundaries may not align perfectly, the resulting meshes cannot be seamlessly merged. Therefore, for whole-hippocampus modeling, the subfield labels were combined and shrink wrapped to merge into a single object.

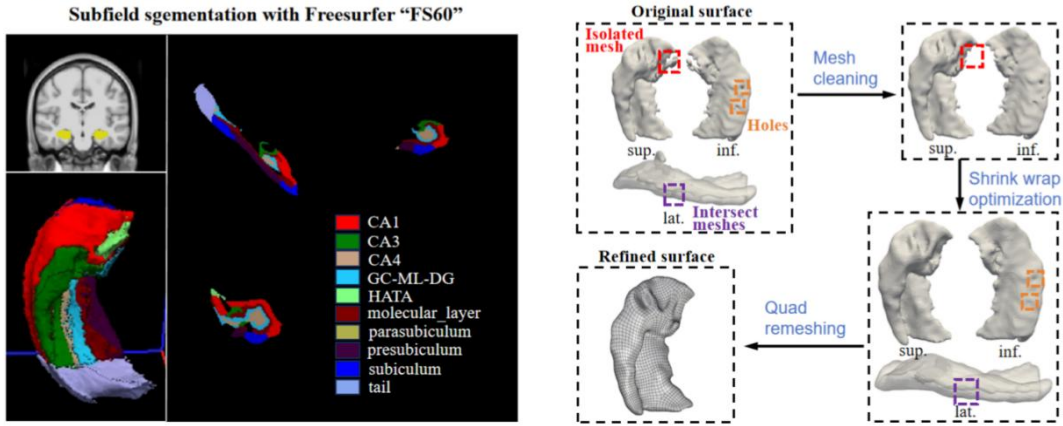

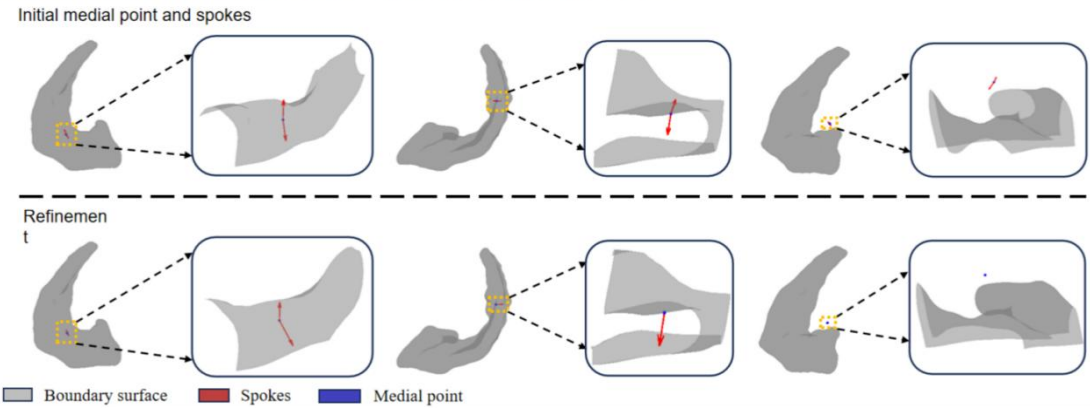

*Figure 3. Geometry-preserving regularization for hippocampal s-rep. A. Surface rigid registration and spatial transformation, mapping an input hippocampus to its initialized s-rep. B. Three cases of spoke optimization: (1) The skeletal point lies inside the boundary and is retained; (2) The skeletal point is outside, and the spoke direction intersects the boundary at two or more points. In this case, the point is reassigned to the first intersection; (3) The skeletal point is outside, and the spoke direction intersects at fewer than two points—the spoke is invalidated and excluded.*

**3.4.2 Spokes refinement**

During template-to-subject deformation, local deviations may occur between the projected sampling points and the actual subfield boundaries. To address this, we refined the subfield-specific spokes using the same optimization strategy described in Section 3.2, which ensures boundary adherence and angular regularity.

For each deformed skeletal point, its spatial relationship to the subfield surface was assessed and categorized into three cases, as illustrated in Fig. 3B:

(1) Valid (inside): If the point lies strictly within the subfield surface, it is retained, and its radial direction is optimized.

(2) Intersecting (outside with ≥2 intersections): If outside the surface but the radial vector intersects the boundary at two or more points, the skeletal point is reassigned to the first intersection, and the radial endpoint to the second.

(3) Invalid (outside with <2 intersections): If the radial vector fails to intersect the surface appropriately (e.g., tangent or disjoint), the spoke is marked invalid.

This refinement step ensures that all spokes used for morphological analysis remain geometrically valid and anatomically consistent across subjects.

For subjects with longitudinal data, a consistent set of spokes initialized at baseline was applied across all time points. During each follow-up observation, only the spoke lengths were re-optimized according to the updated hippocampal and subfield boundaries. This approach enables precise mapping of longitudinal morphological changes, such as atrophy. It is important to note that in some regions, thickness measurements may yield a value of zero. This occurs when atrophy causes the spoke to degenerate to a single point, meaning that the spoke origin and endpoint coincide, which indicates a complete loss of tissue at that location.

### 3.5 Subfield-aware thickness mapping

The global hippocampal morphometric definitions have been previously described in the ARMM framework (Gao et al., 2024b). In brief, global thickness is defined as the sum of the lengths of the superior and inferior spokes at each skeletal point. These can also be reported separately as superior or inferior thickness based on individual spoke lengths. Width is defined as the lateral-skeletal extent measured along the lateral-skeletal axis of the ARMM skeletal surface. Length is the arc length of the long axis traced along the skeletal surface. Curvedness is quantified as the integrated curvature of this long-axis.

For longitudinal data, morphometric changes over time are tracked using subject-specific spoke fields, which are initialized from the baseline and refined on each follow-up scan. Due to structural atrophy, some spokes, particularly superior or inferior spokes, may degenerate to zero length in the follow-up surfaces, reflecting localized thinning or collapse of the structure.

For subfield thickness estimation, let a given subfield contain $N$ skeletal points. The thickness at the $i$-th point is defined as:

$$T_i = \begin{cases} \| pt_{i,up} - ps_i \| + \| pt_{i,down} - ps_i \| \text{ , if both spokes intersect}; \\ \| pt_{i,up} - ps_i^{int} \| \text{ , if only up spoke direction intersect}; \\ \| pt_{i,down} - ps_i^{int} \| \text{ , if only down spoke direction intersect.} \end{cases} \quad (i = 1, \ldots, N)$$

Where
- $ps_i$ is the $i$-th skeletal point;
- $T_i$ is the thickness at the $i$-th skeletal point;
- $ps_{i,up}$ and $ps_{i,down}$ are the updated upper and lower skeletal points after radial axis optimization;
- $pt_{i,up}$ and $pt_{i,down}$ are the corresponding radial endpoints;
- $ps_i^{int}$ is the first intersection point of up/down spoke direction and the subfield boundary.

The definition of local subfield thickness depends directly on the geometric relationship between the skeletal element and the subfield boundary (Fig. 3B). When the skeletal point lies inside the subfield, thickness is computed as the sum of the lengths of both its intersecting spokes. If the point is exterior yet one spoke intersects the boundary at multiple points, thickness is derived from the distance between those intersections. In cases where neither spoke forms a valid intersection with the boundary, the region is considered geometrically degenerate and assigned an invalid thickness value. This approach ensures that only valid geometric contributions are incorporated into the final estimation of subfield thickness.

## 4. Experiment setup

### 4.1 Implementations

All experiments were conducted on a workstation running Ubuntu 18.04 equipped with dual

Intel Xeon Silver 4310 CPUs (48 threads, 2.10 GHz) and an NVIDIA TITAN RTX GPU with 24 GB memory. The HippMetric framework implemented primarily in Python. Boundary surfaces and subfield meshes were processed using VTK, which provides tools for surface extraction, mesh cleaning, quad remeshing, and geometric querying. Template-to-subject transformations were performed with Deformetrica, which supports diffeomorphic shape modeling and enables consistent mapping of skeletal points across individuals.

The template skeleton consisted of 549 skeletal points, each paired with superior and inferior spokes, yielding a total of 1,098 thickness measurements. The hippocampal long-axis was divided into 23 cross-sectional layers to compute 46 width measurements. These resolutions can be adjusted by modifying the density of skeletal points as needed for downstream applications. During surface optimization, target edge lengths ranged from 0.2 to 0.6 mm, and quad counts from approximately 1,000 to 4,000 per subfield, producing smooth and topologically coherent meshes suitable for skeletal refinement.

For diffeomorphic deformation, the noise standard deviation was fixed at 0.05. The Gaussian kernel width was tuned (3–7 mm) in increments of 0.5 to minimize surface mismatch. The smallest kernel width that generated a deformation closely matching the target surface was selected for each subject. In spoke refinement, the step size for updating spoke lengths was set to 0.1 mm with a maximum of 1,000 iterations. Angular adjustments used a step size of 0.1 rad with up to 50 iterations. These parameters provided stable convergence for both cross-sectional and longitudinal modeling.

Surface accuracy was evaluated by comparing the HippMetric-generated RBOS boundaries against surfaces extracted from ITK-SNAP (Yushkevich PA et al., 2006), as well as against reconstructions derived from SPHARM-PDM (Paniagua B et al., 2012), a widely used analytic surface reconstruction method, implemented via the SlicerSALT (Vicory J et al., 2023) graphical interface. Subfield thickness accuracy was assessed against manual measurements as reference.

Reproducibility was examined using repeated scans acquired approximately one year apart, since cognitively normal and pathology-negative individuals are expected to show minimal anatomical change over this interval. To exclude the possibility that measurement agreement stems from differential information encoding across methods, we performed a compactness analysis to verify whether the feature sets from different methods exhibit comparable information density and stability. HippMetric was compared with hippunfold (DeKraker J et al., 2022), a boundary-based morphometry pipeline executed in Docker under the same workstation configuration, with no GPU required and publicly available source code. This comparison evaluates whether the skeletal-representation approach improves robustness to noise and segmentation inconsistencies.

To assess correspondence accuracy in both individual and longitudinal modeling, manually annotated anatomical landmarks were used as ground truth. HippMetric was benchmarked against three representative shape-modeling frameworks, namely, cm-rep (Yushkevich PA et al., 2009), ds-rep (Liu Z et al., 2021), and SPHARM-PDM, to evaluate its ability to produce anatomically coherent pointwise correspondences. cm-rep was compiled from the publicly available C++ implementation, requiring CMake configuration. The ds-rep was obtained from the SHANAPY Python implementation (Liu Z et al., 2023).

Finally, the utility of HippMetric-derived morphometric features in AD applications was assessed by comparing them with traditional volumetric biomarkers, including whole-hippocampal and subfield volumes.

**4.2 Datasets**

We evaluated HippMetric using two independent AD cohorts, including Alzheimer's Disease Neuroimaging Initiative (ADNI) database (Jack CR et al., 2008) and The Greater-Bay-Area Healthy Aging Brain Study (GHABS) (Liu Z et al., 2024). All participants underwent baseline Aβ PET imaging using either 18F-florbetapir (FBP) or 18F-florbetaben (FBB). Aβ status was determined based on Centiloid values, with a threshold of 24.4; scans with values above this threshold were classified as Aβ-positive.

ADNI is a longitudinal multi-site study designed to assess imaging and biomarker changes across the AD spectrum. We selected 465 participants with T1-weighted structural MRI from the ADNI-2, GO, and 3 phases, including both Aβ-positive (A+) and Aβ-negative (A−) individuals. Cognitive status was assessed, and participants were classified as cognitively unimpaired (CU), mild cognitive impairment (MCI), or dementia, with the latter two groups combined as cognitively impaired (CI). Aβ-negative CU individuals served as the healthy control group. All participants underwent baseline scans, and a subset (n = 201) had at least one follow-up MRI and PET scan.

GHABS is an ongoing community-based study conducted in Southern China. We included 362 participants with Aβ PET at baseline, of whom 132 individuals had longitudinal T1-weighted MRI. The cognitive status was also determined for individuals with CU, MCI, and dementia, enabling validation of HippMetric in a non-Western population.

Based on longitudinal data from both cohorts, participants were further classified as converters or non-converters. Converters included individuals who progressed from CU to MCI or dementia, or from MCI to dementia, while non-converters were those whose cognitive status remained stable over time (stable CU or MCI). Key demographic characteristics are summarized in Table 1. All MRI data were collected using 3.0 T scanners with an isotropic resolution of 1 mm.

We used different subsets of the ADNI and GHABS datasets for each experimental component. To evaluate thickness accuracy, we selected subjects from the ADNI cross-sectional dataset and manually measured subfield thickness for comparison. Reproducibility and sensitivity to biological differences of morphometric measurements was assessed using longitudinal data from both ADNI and GHABS. Cross-sectional correspondence accuracy was evaluated using the ADNI cross-sectional dataset with manually annotated landmarks. Validation of HippMetric measurements was conducted using data from the ADNI and GHABS cohorts. Cross-sectional data were used for classification, while longitudinal data from converters and non-converters within these cohorts were used for prediction analysis.

*Table 1. Demographic and diagnostic characteristics of the included cohorts.*

|  | ADNI | | GHABS | |
| --- | --- | --- | --- | --- |
| *827 paticipants with Aβ PET at baseline* | | | | |
| Diagnose | A+/CI | A-/CU | A+/CI | A-/CU |
| No. patients | 183 | 282 | 88 | 274 |
| Age, years | 75.8 ± 7.9* | 71.1 ± 6.5* | 70 ± 9.5* | 64.8 ± 7.6* |
| Female, no. (%) | 82 (44.8%)* | 167 (59.2%)* | 56 (63.6%) | 163 (59.5%) |
| Education, years | 15.9 ± 2.5* | 16.8 ± 2.3* | 3.7 ± 1.6* | 5.5 ± 1.1* |
| APOE ε4 carrier, no. (%) | 116 (63.4%)* | 72 (25.5%)* | 49 (55.7%)* | 67 (24.5%)* |
| MMSE | 26.5 ± 3.1* | 29.2 ± 1.0* | 19.7 ± 6.3* | 28.5 ± 1.6* |
| CDR | 0.53 ± 0.21* | 0.03 ± 0.12* | 0.94 ± 0.7* | 0.03 ± 0.13* |

***303 participants with cross-sectional and longitudinal Aβ PET scans***

| | | | | |
|---|---|---|---|---|
| Baseline Diagnose | A+/CI | A-/CU | A+/CI | A-/CU |
| No. patients (%) | 18 | 183 | 36 | 96 |
| Age, years | 79.4 ± 6.7* | 71.8 ± 6.1* | 69.0 ± 8.8* | 63.7 ± 8.0* |
| Female, no. (%) | 6 (33.3%)* | 107 (58.5%)* | 19 (52.8%)* | 55 (57.3%)* |
| Visits of MRI (points) | 3.5 ± 1 | 2.8 ± 0.8 | 2.2 ± 0.4 | 2.3 ± 0.5 |
| Duration of MRI (years) | 3.2 ± 1 | 3.4 ± 1.2 | 1.6 ± 0.7 | 1.6 ± 0.7 |

***382 participants with either stable cognition (CU or MCI-stable) or clinical progression (CU →MCI/AD, MCI →AD)***

| | | | | |
|---|---|---|---|---|
| Diagnose | convertor | non-convertor | Converter | Non_Converter |
| No. patients (%) | 66 | 265 | 6 | 45 |
| Age, years | 74.8 ± 6.6* | 73.0 ± 7.2* | 66.7 ± 4.8* | 63.3 ± 6.3* |
| Female, no. (%) | 31 (47%)* | 138 (52.1%)* | 4 (66.7%)* | 28 (62.2%)* |
| Baseline Diagnosis | CU:33 MCI:33 | CU:172 MCI:93 | CU:5 MCI:1 | CU:40 MCI:5 |
| Visits of MRI (points) | 3.2 ± 1 | 2.9 ± 0.9 | 2.0 ± 0.0 | 2.3 ± 0.5 |
| Duration of MRI (years) | 3.3 ± 1.4 | 2.9 ± 1.3 | 1.7 ± 0.7 | 1.8 ± 0.6 |

*Note. Data are presented as the number of participants (no.) and percentage (%) or median and interquartile range (IQR). Continuous variables (age, education years, MMSE, and CDR scores) were compared using independent two-sample t-tests. Categorical variables (sex and APOE ε4 carrier status) were compared using chi-square tests. Benjamini-Hochberg false discovery rate (FDR) was used for group-level multiple corrections (significance level, P < 0.05).*

**4.3 Validation of HippMetric template accuracy**

To assess the performance of the proposed template construction method, we conducted validation from two complementary perspectives: the morphological fidelity of reconstructed subfield surfaces and the precision of thickness measurements derived from their s-reps.

First, we validated the morphological accuracy of RBOS-generated hippocampal subfield surfaces. Since the method adjusts initially free and self-intersecting meshes to match true hippocampal anatomy, our validation emphasized overall shape fidelity.

Three shape similarity metrics were employed for quantitative validation: average surface distance ($Q$), maximum surface distance ($q$), and Hausdorff distance ($HD$). Let $S$ represents the ground truth surface, $T$ denotes the test surface, $p$ and $q$ are points on surfaces $T$ and $S$ respectively, and $n$ is the number of points on the surface $T$. The average surface distance is defined as:

$$Q = \frac{1}{n}\sqrt{\sum_{i=1}^{n} \| p_i - q_i \|^2}, \ i = 1, \ldots, n.$$

The maximum surface distance is defined as:

$$q = max \| p_i - q_i \|, \ i = 1, \ldots, n$$

The Hausdorff distance is defined as:

$$HD(\vartheta, \xi) = max\big(h(\vartheta, \xi), h(\xi, \vartheta)\big)$$

These metrics provide complementary assessments of surface agreement. $Q$ captures global deviation characteristics, $q$ identifies localized maximum point-wise errors, and $HD$ quantifies topological consistency through bidirectional extreme value analysis. Each subfield surface generated by different methods was evaluated against the ground truth surfaces extracted directly from MRI segmentation labels to determine boundary representation accuracy.

Secondly, we validated the measurement accuracy of the FS60 hippocampal subfield thickness template using manual measurements on segmented labels as the ground truth. For manual measurements, thickness was defined as the distance from the superior to the inferior boundary surface at each point, reflecting its direct anatomical interpretation.

**4.4 Validation of HippMetric measurement accuracy, reliability, and correspondence stability**

To assess the accuracy of HippoMetric measurements, this study employed two validation approaches: (1) comparison against manual measurements as the ground truth, (2) evaluation of measurement reproducibility through comparative analysis with Hippunfold, (3) assessment of spatial correspondence accuracy for locally-derived features across cross-sectional and longitudinally, and (4) evaluation of measurement stability under pathological conditions, including severely atrophic hippocampi.

First, to evaluate the accuracy of subfield-level thickness measurements, we performed a validation study using 10 randomly selected individuals from the ADNI dataset. Participants were matched for age, sex, and MMSE scores, with no significant differences across the sample.

Manual thickness measurements derived from Freesurfer-generated label images served as the reference standard. Because the skeletal points defined in the FS60 template could not be precisely localized on voxel-wise labels, we manually sampled an equal number of points for each individual and subfield. The mean of these manual measurements was taken as the reference thickness. The subfield-level mean thickness error was then calculated as:

$$D = \frac{\sum_{i=1}^{n} TF_i}{n} - \frac{\sum_{i=1}^{n} TM_i}{n},$$

where $n$ is the number of valid skeletal points (also the number of manual measurements), $TF$ is the thickness measured by HippMetric, $TM$ is the corresponding manual measurement, $D$ denotes the mean subfield thickness error. For subfields with composite skeletons (e.g., CA1, GC-ML-DG, molecular_layer), errors were computed separately for each anatomical portion.

Measurement reproducibility was quantified using the intraclass correlation coefficient (ICC), which was selected for its ability to evaluate both consistency and absolute agreement between repeated measurements. The ICC was calculated using a two-way mixed-effects model (ICC(3,1)), based on the formula:

$$ICC(3,1) = \frac{MS_R - MS_E}{MS_R + (k-1)MS_E},$$

where $MS_R$ is the mean square for rows (subjects), $MS_E$ is the residual mean square, and $k$ is the number of measurements per subject.

To complement the reproducibility assessment and evaluate the information efficiency of extracted features, we performed principal component analysis to assess feature compactness. Compactness reflects how efficiently a feature set captures the underlying variance in the data, with higher compactness indicating that fewer features are needed to represent the same information. Comparable information density and stability among feature sets in different methods can confirm that the observed reproducibility reflects methodological reliability rather than incidental differences in shape representation.

The cross-subject and longitudinal correspondence accuracy is quantified by computing point-wise alignment errors between algorithm-derived correspondences and manually annotated landmarks on hippocampal surfaces reconstructed using different shape representation methods. This evaluation serves as a proxy for assessing subfield-level correspondence reliability, since the accuracy of subfield-level morphometric analysis critically depends on the point-wise correspondences established at the whole-hippocampal level.

We evaluated the stability of HippMetric under pathological conditions across clinically relevant diagnostic groups by quantifying reconstruction accuracy on hippocampal surface derived from cognitively unimpaired (CU) and cognitively impaired (CI) subjects. Geometric fidelity was assessed using established shape metrics, including mean and maximum surface distance (Q and q), Hausdorff distance (HD), surface area error (Ae), and curvature error (Ce), to comprehensively characterize reconstruction performance across disease stages. The curvedness at each vertex is defined by

$$c=\sqrt{(k_{max}^2 + k_{min}^2)/2},$$

where $k_{max}$ and $k_{min}$ represent the main curvature.

**4.5 Validation of HippMetric measurements for classification and longitudinal prediction in AD**

To assess the clinical relevance of HippMetric-derived hippocampal substructural features, we examined these features' cross-sectional discriminative power and longitudinal prognostic value. Cross-sectional analyses evaluated the ability of these features to differentiate amyloid-negative cognitively unimpaired (A−/CU) individuals from amyloid-positive cognitively impaired (A+/CI) participants at baseline. Longitudinal analyses tested whether the same features could predict future clinical conversion during follow-up, compared with conventional volumetric measures. Two independent cohorts were used: the ADNI dataset served as the discovery cohort for model development and feature selection, and the GHABS cohort provided external validation.

For the cross-sectional evaluation, we implemented a multi-stage feature selection procedure within the ADNI cohort. Features were sequentially retained if they demonstrated: (i) significant group differences (Cohen's d, FDR-corrected p < 0.05) assessed using w-score normalization with age, sex, and education as covariates, followed by permutation testing with FDR correction; (ii) significant positive correlations with cognitive decline measured by Pearson correlation coefficients; and (iii) positive importance scores in random forest classification. The selected features were subsequently validated in the independent GHABS cohort through bidirectional cross-testing and internal cross-validation. Classification performance was quantified using ROC analysis across different feature sets, including both traditional volumetric measures and Hippometric measurements.

For longitudinal prediction, the features identified from the cross-sectional analysis were used to train a regularized logistic regression model to identify individuals who would convert to a more severe clinical stage. The model employed an elastic net penalty ($\alpha$ = 0.5) with the regularization parameter $\lambda$ optimized via 5-fold cross-validation, and its generalizability was further assessed in the GHABS cohort. Predictive accuracy was evaluated using the AUC, and prognostic utility was assessed via Kaplan-Meier survival analysis with log-rank tests. For risk stratification, the GHABS cohort was dichotomized at the median of model-derived risk probabilities, classifying individuals with scores above the median as high-risk and those below as low-risk.

## 5. Results

**5.1 Template accuracy**

The evaluation was performed on 40 T1-weighted MRI scans (20 MCI, 20 CN). We compare our RBOS against SPHARM-PDM and ITK-SNAP using voxel-based surfaces reconstructed directly from T1w segmentation masks as the ground truth. The results are shown in the top left panel of Fig. 4 (Table. S1), ITK-SNAP achieved the lowest surface distance errors, followed closely by RBOS, while SPHARM-PDM yielded the highest errors, especially in geometrically complex regions. Despite slightly lower numerical errors, ITK-SNAP produced surfaces with disconnected or self-intersecting components (see the top right panel of the Fig. 4). In contrast, RBOS produced clean, watertight surfaces with sub-millimeter accuracy, outperforming SPHARM-PDM and providing more anatomically valid reconstructions.

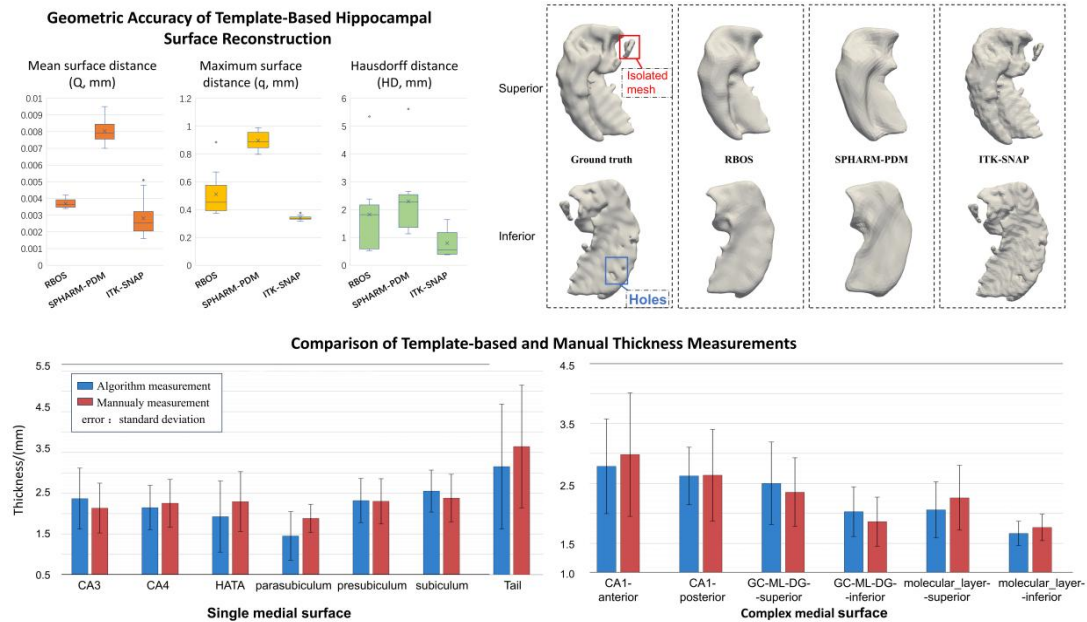

*Figure 4.* A Thickness measurement error compared with manual measurements. B. and C. ICC distributions of subfield measurements from HippMetric and Hippunfold.

In the assessment of thickness measurement accuracy, among subfields with single skeletons, the largest deviation was observed in the tail (Q = 0.5 mm), while CA3, CA4, and subiculum showed errors below 0.25 mm. For composite subfields, the maximum error across the six regions was 0.20 mm. As shown in bottom panel of Fig. 4, thickness errors in all subfields were well below the 3T T1w resolution (1mm).

**5.2 Measurement accuracy and stability**

To assess measurement reproducibility, we used longitudinal MRI data from A-/CU subjects in ADNI and GHABS, including 90 individuals with two scans acquired within 12 months; all morphological features were extracted at both time points for reliability analysis. Results are summarized in Fig 5A, with detailed values reported in Table S2. The largest mean error was observed in the GC-ML-DG inferior subfield (1.20 mm). All other subfields exhibited mean errors ranging from 0.4 mm to 0.8 mm, which is smaller than the native voxel size of 3T T1w MRI (1 mm).

Test-retest reliability of HippMetric measurements was quantified using the intraclass correlation coefficient (ICC) analysis, where values closer to 1 indicate stronger measurement stability. The analysis evaluated both local thickness measurement stability (local ICC and variance ICC) and global reliability (Mean ICC and Global ICC). The numerical distribution and spatial patterns of ICC values across all subfield points are visualized in Fig. 5B, and detailed values are provided in Tables S3 and S4. To benchmark subfield morphometric reproducibility, we compared HippMetric against HippUnfold, currently the most advanced hippocampal subfield morphometry method, using the same dataset.

Results demonstrated that HippMetric's Inf_thickness and Sup_thickness (representing inferior and superior hippocampal thickness, respectively), as well as right-tail thickness, showed ICC values around 0.6, comparable to those of Hippunfold. In contrast, HippMetric's width-related features, including lateral width (Lat_width), medial width (Med_width), and lateral-medial width (width), as well as length, achieved mean ICC values above 0.91. Among subfield thickness measures, CA1, CA4, HATA, parasubiculum, presubiculum, and tail reached mean ICC values above 0.2, performing comparably to Hippunfold's curvature features. Overall, HippMetric exhibited smaller standard deviations in local ICC values than Hippunfold. The maximum variation in HippMetric was observed in the molecular_layer thickness (0.0415), which was still lower than the minimum variation in Hippunfold (0.048, for hipp_thickness).

Spatial ICC maps revealed that HippMetric's local ICC values were more clustered, with higher reliability along the mid-region of the hippocampal long axis, indicating superior stability in these regions. Lower ICC values were observed in medial-lateral regions and the anterior portion of the hippocampal head. In comparison, Hippunfold's ICC distribution appeared more heterogeneous, resembling a salt-and-pepper pattern, with generally higher reliability in the posterior hippocampus and medial aspect of the head, and lower reliability in the anterior head region.

The results of global ICCs are listed in Supplementary Material TableS2 and S3. HippMetric's width-related features, Length, and left CA1 thickness all exceeded 0.9, whereas Hippunfold only achieved values above 0.9 for hipp_surface and gyrification. HippMetric's Inf_thickness, Sup_thickness, CA3, CA4, para_sub, pre_sub, and tail all reached ICC values above 0.7, comparable to Hippunfold's hipp_thickness.

Results of compactness showed comparable compactness between the two methods: the first 10 principal components explained 61.2% and 67.0% of the total variance for HippMetric and HippUnfold, respectively. To reach 90% explained variance, HippMetric required 21 principal components while HippUnfold required 19. This similarity in compactness indicates that both methods extract comparable amounts of morphological information from hippocampal MRI.

However, a critical difference emerged in feature efficiency. HippMetric achieved this information level using only 1,849 features, while HippUnfold required 14,524 features, 7.85 times more than HippMetric. Consequently, HippMetric's feature efficiency (variance explained per feature $\times 10^3$) was 33.080, compared to 4.615 for HippUnfold, representing a 7.2-fold improvement. The compactness analysis complements the ICC results by demonstrating that while both methods extract comparable amounts of morphological information, HippMetric achieves this with 87% fewer features. This improvement in feature efficiency, combined with its clustered ICC spatial patterns, indicates that HippMetric maintains measurement reliability while providing a more efficient feature representation suitable for longitudinal biomarker discovery.

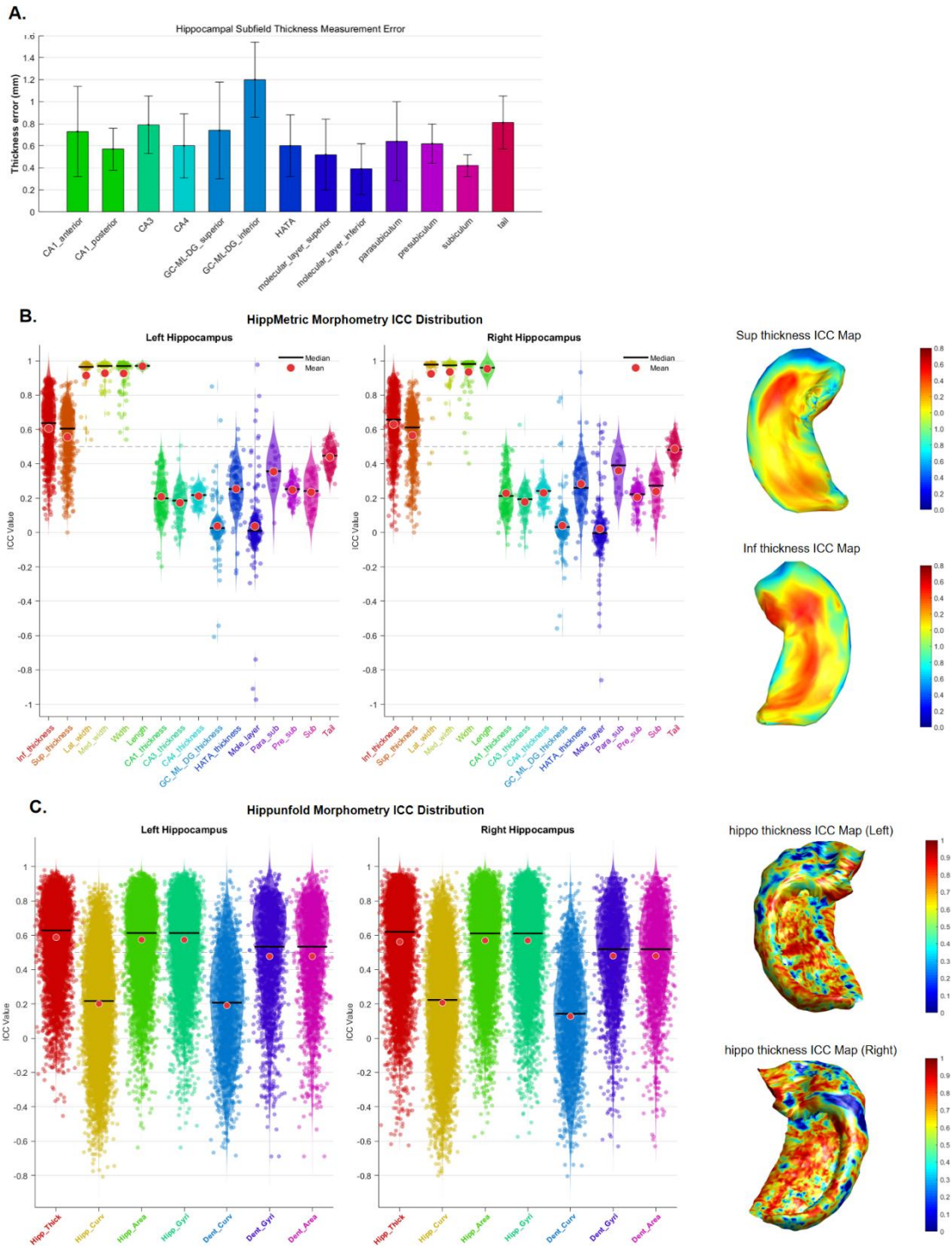

**Figure 5.** *A Thickness measurement error compared with manual measurements. B. and C. ICC distributions of subfield measurements from HippMetric and Hippunfold.*

## 5.3 Cross-subject and Longitudinal correspondence performance

We first evaluated the accuracy of cross-subject correspondence. Fig. 6A shows representative three hippocampal surfaces reconstructed by HippMetric, SPHARM-PDM, cm-rep, and ds-rep, compared with ground-truth surfaces extracted from anatomical labels. While all methods produced anatomically plausible reconstructions, HippMetric and SPHARM-PDM exhibited higher geometric fidelity to the original surfaces, as indicated by the similarity of curvature maps. In contrast, ds-rep showed substantial deviation from ground-truth due to its reduced surface resolution (low mesh vertexes).

To evaluate alignment accuracy, we first identified all eligible left-hemisphere hippocampal scans from ADNI (n = 465). Because manual landmark is time-consuming and requiring high intra-rater consistency, a representative set of 30 subjects was randomly selected to cover the range of inter-individual shape variability. For each selected case, five anatomically identifiable landmarks were manually placed on both the ground-truth and reconstructed surfaces for error quantification. Across all methods and subjects, this resulted in 600 annotated points. The left panel of Fig. 6A illustrates the spatial layout of these landmarks, including the anterior and posterior extremities, medial folding apex, and lateral curvature center.

Alignment errors were quantified as Euclidean distances between algorithm-generated and manually annotated points. As summarized in the right panel of Fig. 6A, HippMetric achieved the lowest average correspondence errors across all landmarks, with a maximum mean error of 1.66 mm at the anterior folding apex. Errors at other locations remained below 1.1 mm on average, with standard deviations of less than 1 mm. In comparison, cm-rep and ds-rep showed moderate alignment accuracy, with maximum errors reaching 4.64 mm and 3.41 mm, respectively. SPHARM-PDM exhibited the largest variability, with a maximum error of 9.97 mm, likely due to pole flipping. Detailed correspondence errors for each method across all landmark locations are reported in Table S5.

We then conducted a simulation study to compare the longitudinal alignment performance across different methods. We randomly selected 15 baseline hippocampi from the ADNI dataset and manually simulated progressive atrophy by iteratively eroding the anterior region, generating three follow-up surfaces (t1-t3) with increasing severity of atrophy. The posterior region was preserved to isolate the atrophic effects. All surfaces were reconstructed using HippMetric, SPHARM-PDM, cm-rep, and ds-rep. For evaluation, three landmarks were manually placed within the atrophic region on each surface (see Fig. 6B), and correspondence errors were computed as Euclidean distances between algorithm-predicted and manual landmark positions.

Visual inspection showed that HippMetric preserved anatomically coherent point-wise trajectories that closely followed the atrophic process. Competing methods exhibited non-local displacements from global reparameterization, often shifting landmarks outside the affected region. SPHARM-PDM and cm-rep, in particular, could not confine deformation to the atrophic zone, resulting in anatomically implausible drift. By contrast, HippMetric maintained spatial specificity and captured meaningful surface collapse at advanced atrophy stages, reflected by near-zero spoke lengths, an effect not achieved by other models.

Quantitatively, HippMetric showed the smallest correspondence errors across all subjects and time points (right panel of Fig. 6B). The mean error was 1.68 mm across the four longitudinal time points, with a maximum error of 2.81 mm at time point t3 (location index 3). In comparison, cm-rep exhibited a mean error of 10.57 mm and a maximum of 20.48 mm at location 2; ds-rep showed a mean error of 4.65 mm and a maximum of 20.07 mm at location 3; and SPHARM-PDM showed a mean error of 3.56 mm and a maximum of 16.87 mm. While correspondence errors increased over time for HippMetric, ds-rep, and SPHARM-PDM, the latter two exhibited substantially larger drifts, particularly at time point t3. Table S6 lists the correspondence errors of all methods for each landmark across all longitudinal time points.

## A. Cross-sectional correspondence accuracy

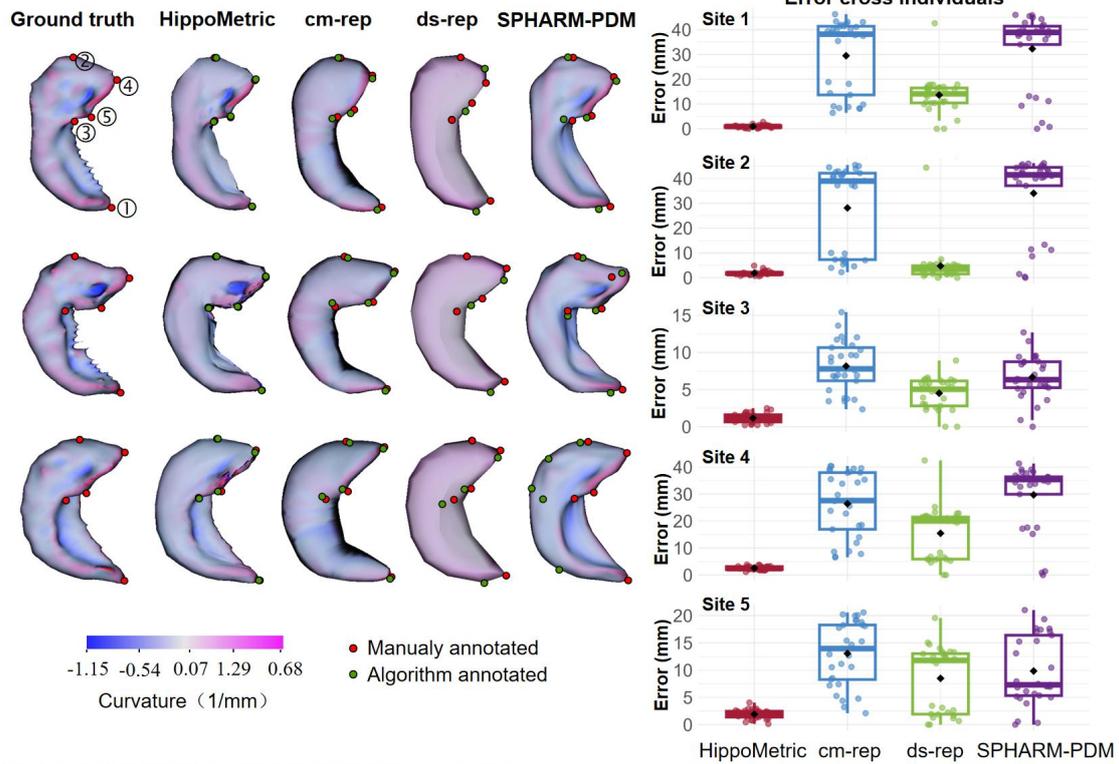

## B. Longitudinal correspondence accuracy

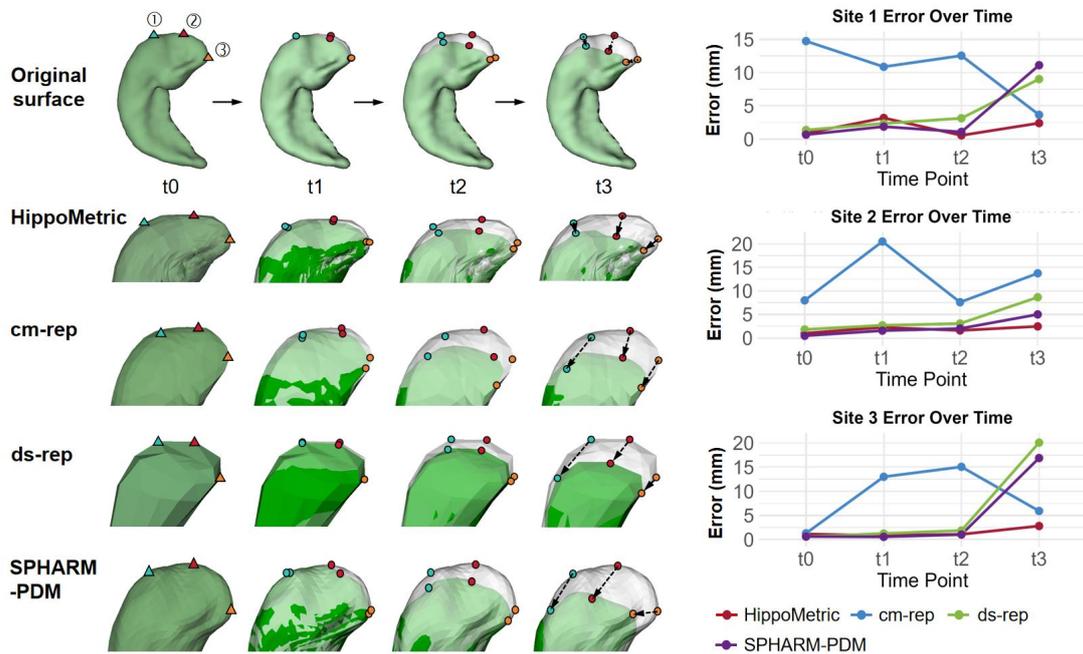

***Figure 6.*** *Qualitative and quantitative evaluation of cross-sectional and longitudinal morphological correspondence errors. A. Left: curvature maps illustrating local cross-sectional correspondences for three randomly selected hippocampal samples, with five representative locations marked for correspondence error evaluation; Right: comparison of local cross-sectional correspondence errors across different hippocampal shape representation methods. B. Left: longitudinal local correspondence patterns during a simulated hippocampal atrophy process, along with morphological reconstructions and correspondence mappings generated by different*

*methods; Right: comparison of longitudinal correspondence errors among methods in capturing morphological progression.*

**5.4 Sensitivity to biological differences**

In AD, cognitive impairment (CI) is typically accompanied by more pronounced hippocampal atrophy compared with cognitively unimpaired (CU) individuals, providing a useful context for evaluating whether reconstruction accuracy declines with increasing biological severity. In this analysis, we used the combined ADNI and GHABS cohorts, comprising 556 A﹣/CU participants and 271 A+/CI participants, ensuring sufficient representation across the two biological conditions. As summarized in Table 2, HippMetric maintained stable accuracy across CU and CI groups, with no significant differences ($p>0.05$) in all error metrics (Q, q and HD) across diagnostic categories. In contrast, SPHARM-PDM and cm-rep showed progressively larger errors with increasing pathology severity, though not significant. Statistical testing confirmed that HippMetric's reconstruction accuracy was unaffected by biological condition, demonstrating robustness to atrophic shape change.

*Table 2. Errors between the hippocampal shape reconstructed by different methods and the ground truth, comparing disease and control groups*

| Group - Surface error | HippMetric | cm-rep | ds-rep | SPHARM-PDM |
|---|---|---|---|---|
| CU-Q (mm) | 0.003±0.001 | 0.039±0.022 | 0.144±0.108 | 0.013±0.001 |
| CU -q (mm) | 0.738±0.016 | 1.714±1.723 | 9.386±9.934 | 1.067±0.161 |
| CU -HD (mm) | 1.489±0.301 | 1.489±0.301 | 11.265±8.491 | 14.890±10.073 |
| CU-Ae (mm$^2$) | 0.887±0.408 | 11.933±13.974 | 28.781±14.997 | 1.717±0.439 |
| CU-Ce (1/mm) | 0.067±0.069 | 0.072±0.082 | 0.137±0.049 | 0.061±0.042 |
| CI-Q (mm) | 0.003±0.001 | 0.069±0.210 | 0.119±0.102 | 0.014±0.001 |
| CI-q (mm) | 0.759±0.076 | 8.686±4.018 | 7.203±9.321 | 1.076±0.165 |
| CI-HD (mm) | 1.809±0.388 | 25.035±9.868 | 9.396±8.043 | 17.641±9.332 |
| CI-Ae (mm$^2$) | 0.883±0.453 | 13.399±15.030 | 32.121±18.064 | 1.745±0.548 |
| CI-Ce (1/mm) | 0.057±0.091 | 0.166±0.936 | 0.088±0.301 | 0.072±0.094 |

**5.5 Diagnostic and prognostic performance of hippocampal substructural features in AD and cognitive impairment**

Cross-sectional analysis of the ADNI cohort revealed morphological differences in hippocampal substructures between A-/CU and A+/CI groups. Among 1,096 features examined, 521 showed significant group differences (235 left, 286 right hippocampus). Width measurements demonstrated the most substantial alterations, with 150 features (80.6% of all width features) showing significant differences. Reductions were observed in hippocampal head length, tail length, and total length. Thickness alterations were present in 368 features (40.6% of thickness features).

Analysis in the GHABS cohort identified 411 features with consistent group differences across both datasets, representing 78.9% of ADNI significant features. Fig. 7A (top-left) shows the top 10 features by effect size (Cohen's d) in ADNI with corresponding values in GHABS. The most pronounced differences were localized to posterior hippocampal dimensions, particularly mediolateral width at the 8th lamella of the hippocampal tail.

We examined the relationship between significant features and cognitive decline rates. In

ADNI, 383 features showed positive correlations with cognitive deterioration, while one feature (Left mole_layer.20) showed a negative correlation. In GHABS, 221 features maintained positive correlations, two showed negative correlations, and 160 were non-significant. Figure 7A (top-right) displays the top cognition-associated features, with tail dimensions showing consistent associations.

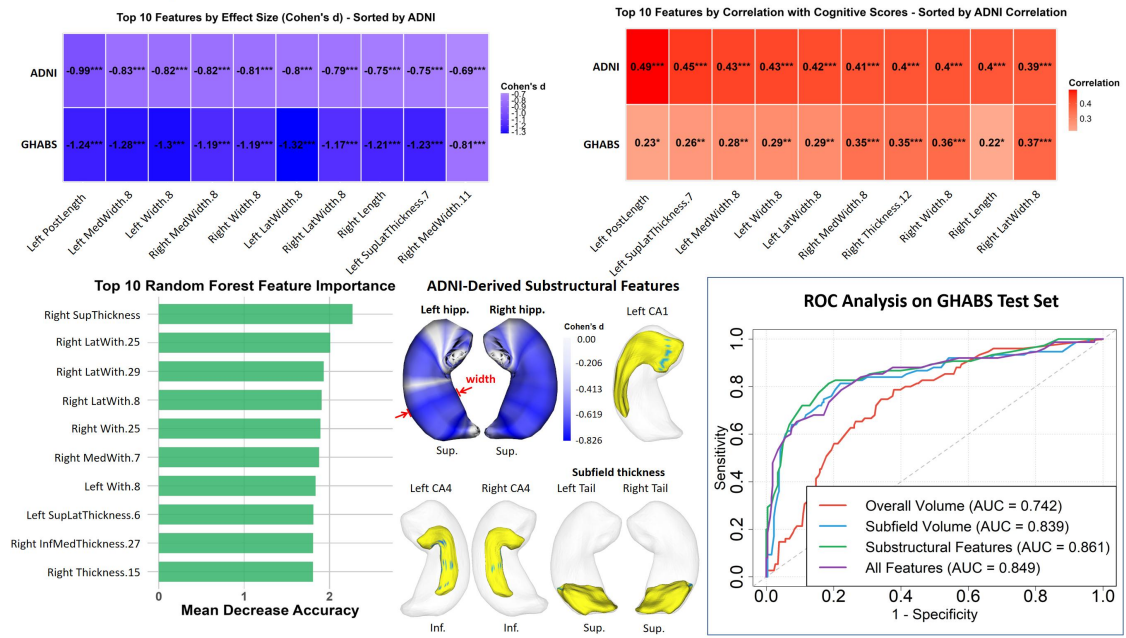

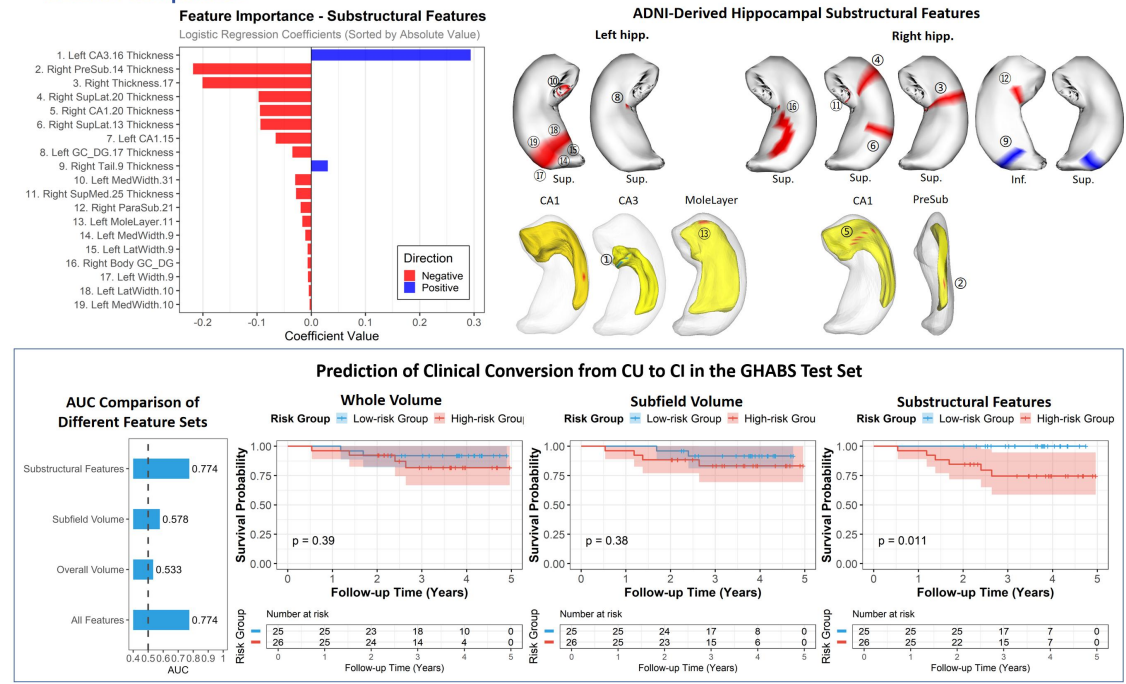

*Figure 7.* Hippocampal substructure by HippMetric versus traditional volumetric features for cross-sectional classification and longitudinal conversion prediction.

Random forest classification for A-/CU vs A+/CI discrimination in ADNI achieved an AUC of

0.735. The selected features yielded an AUC of 0.861 in GHABS, compared to 0.742 for overall hippocampal volume, 0.839 for subfield volumes, and 0.849 for all combined features.

In longitudinal prediction of clinical conversion, regularized logistic regression identified 19 predictors from 1,096 initial features for conversion from CU to CI (Fig. 7B, first row). Key features included left CA3 thickness at lamella 16 (coefficient = 0.294), right presubiculum thickness at lamella 14 (coefficient = -0.218) and right hippocampal thickness at lamella 17 (coefficient = -0.201).

In GHABS longitudinal validation, the substructural feature model showed an AUC of 0.774, compared to 0.533 for overall hippocampal volume and 0.578 for subfield volumes. Among 51 participants, the model classified 26 as high-risk with 6 conversions (23.1%), while no conversions occurred among 25 low-risk individuals (log-rank p = 0.011).

### 5.6 Modularity and runtime performance

To evaluate the computational efficiency of HippMetric, we compared its runtime and workflow organization with existing hippocampal morphometry pipelines. On our workstation, the full pipeline (excluding hippocampal subfield segmentation) requires approximately 7 minutes per subject. Batch processing across multiple subjects demonstrated linear scalability with negligible memory overhead.

HippMetric adopts a modular architecture that allows each processing step to be executed independently, including FreeSurfer-based subfield segmentation, surface extraction, surface refinement, surface registration, spatial transformation, spoke refinement, and morphometric measurement. This design facilitates flexible integration with external tools and efficient execution in large-scale or longitudinal studies.

To further assess its utility, we compared HippMetric with widely used shape modeling and measurement frameworks, including SPHARM-PDM, ds-rep, and hippounfold. The comparison focused on practical aspects, including input-type compatibility, FreeSurfer integration, subfield-level measurement support, modularity, batch-processing capability, longitudinal consistency, anatomical correspondence, supported morphometrics, runtime, failure cases, and output formats.

As summarized in Table 3, HippMetric offers end-to-end processing, spanning from segmentation to subfield-level morphometric quantification, with GPU-accelerated computation and a fully modular workflow, enabling efficient cross-subject and longitudinal analyses. Unlike SPHARM-PDM and ds-rep, which lack subfield-level resolution and can suffer from unstable correspondences or restrictive topology assumptions, HippMetric produces anatomically coherent measurements grounded in lamellar and medial-axis organization. Compared with other tools, it offers seamless integration with FreeSurfer, supports multiple input modalities (T1w, label maps, and surfaces), ensures robust longitudinal consistency, and maintains competitive runtime (~7 minutes per hippocampus). Outputs follow standard formats such as CSV, facilitating downstream statistical analysis.

*Table 3. Comparative evaluation of different shape processing pipelines*

| Feature / Tool | HippMetric | ARMM | SPHARM-PDM | ds-rep | hippunfold |
|---|---|---|---|---|---|
| Input type compatibility | **T1w, label, surface** | label, surface | label, surface | SPHARM-PDM based Surface | T1w (BIDS) |
| FreeSurfer integration | √ | × | × | × | × |
| Subfield-level measurement | √ | × | × | × | √ |
| Modularity & batch support | √ | √ | √ | √ | √ |
| Longitudinal consistency | √ | √ | × | × | × |
| Anatomical correspondence | **Medial-axis and lamellar organization based** | Medial-axis and lamellar organization based | spherical based | Medial-axis based | Laplace's equation based |
| Morphometrics supported | **Whole shape: thickness, length, width, curvature; Subfields: thickness** | Whole shape: thickness, length, width | Whole shape: point location | Whole shape: spokes length, skeletal points location | Subfields: thickness, curvature, gyrification |
| Runtime (per hippocampus) | **~7 min (GPU)** | ~5 min (GPU) | ~5 min | ~3min | ~30 min (including segmentation procedure) |
| Failure cases | **Missing values in complex regions** | Rare | Pole flipping | Rare | Mesh unfolding error |
| Output format | **CSV** | CSV | VTK | VTK | GIfTI |

## 6. Discussion and Conclusion

In this study, we developed HippMetric, a skeletal-representation-based framework for hippocampal subfield morphometry, designed to provide a comprehensive set of multi-scale structural metrics with important potential as novel imaging biomarkers of neurodegenerative disease. The framework consists of two major procedures: manual template construction and automated morphometric extraction for any individual hippocampus. The manually built template is designed to accommodate different subfield segmentation protocols; in this work, the template was tailored to the FreeSurfer FS60 atlas and can be seamlessly integrated into the FreeSurfer hippocampal subfield segmentation workflow. The automated pipeline includes surface refinement, spatial deformation, spoke refinement, and computation of subfield-wise geometric features. All these steps are encapsulated into a unified pipeline, allowing fully automated processing with minimal user intervention.

A key advantage of HippMetric lies in its reliance on s-reps, which provide strong geometric stability and consistent cross-subject correspondence because the representation is rooted in the intrinsic skeletal structure of 3D objects. This intrinsic skeletal geometry offers a principled coordinate system for capturing local shape, allowing correspondences to be established in a way that is less affected by small perturbations on the boundary surface. To enhance the biological correspondences, in this work we bridge the skeletal geometry with hippocampal lamellae axis that closely tied to hippocampal functional organization. This skeleton-based formulation offers distinct advantages in longitudinal analysis, where conventional surface-dependent pipelines often conflate genuine atrophy with artifactual changes introduced by scan-rescan noise or segmentation inconsistency. By anchoring correspondences to the stable internal skeletal scaffold and enforcing geometrically consistent constraints, HippMetric substantially reduces sensitivity to such technical confounds. As a result, it better preserves the temporal continuity of shape evolution and more reliably distinguish genuine biological atrophy from artifacts arising from surface noise or alignment inconsistencies.

We conducted a series of experiments to evaluate both the accuracy of the constructed template and the reliability of automated measurements. Template accuracy was validated through comparisons of boundary surfaces with conventional surface reconstruction methods, and skeletal thickness measures were benchmarked against manual annotations. The automated measurement pipeline was further evaluated across four aspects: (1) Accuracy of subfield-wise thickness, compared with manual measurements; (2) Local and global reproducibility, benchmarked against hippunfold; (3) Cross-subject and longitudinal correspondence accuracy, compared with widely used shape models (cm-rep, ds-rep, and SPHARM-PDM); (4) Robustness to pathological variation, such as highly atrophic hippocampi in AD.

Across these evaluations, HippMetric achieved measurement errors of less than 0.5 mm relative to manual ground truth, demonstrated superior stability compared to Hippunfold, and provided more accurate correspondence than existing shape models. Notably, correspondence stability along longitudinal trajectories showed substantially smaller errors (mean 1.68 mm compared with 3.56 to 10.57 mm in existing models), indicating that HippMetric more faithfully captures spatially localized shape changes over time. This level of temporal fidelity is generally not achieved by current boundary-based or point-distribution models.

Another strength of HippMetric is its ability to derive a new class of geometric and anatomically interpretable imaging metrics. Analogous to cortical thickness, we compute curved hippocampal cortical thickness, along with global hippocampal shape descriptors such as overall thickness and width. Experiments on AD datasets revealed that global hippocampal thickness and width serve as strong discriminators between individuals with AD and those who are cognitively normal, suggesting that neurodegeneration involves not only focal subfield changes but also coordinated atrophy across adjacent subfields. Skeletal-lateral narrowing, which is difficult to quantify in traditional models, was consistently observed along the AD continuum. Longitudinal analyses further highlighted CA1, presubiculum, and the molecular layer as key subfields undergoing progressive degeneration, consistent with extensive prior literature. Importantly, subfield-level morphometric changes derived from HippMetric outperformed both whole-hippocampus and subfield-volume features in predicting clinical progression to dementia, underscoring their potential as sensitive imaging biomarkers.

Despite these strengths, HippMetric has several limitations. First, skeletal modeling cannot be reliably applied to every subfield in every individual hippocampus. Substantial inter-subject shape variability can cause diffeomorphic registration to fail, a limitation inherent to current classical deformation methods. Second, some subfields contain small regions that are difficult to measure due to insufficient spoke sampling density. While increasing spoke density can alleviate this issue, it may introduce unrealistic geometry in low-resolution MRI; therefore, the optimal sampling strategy remains application-dependent.

To address these limitations, future work will focus on improving the spatial deformation component, potentially integrating deep learning–based generative models to construct more flexible and robust s-reps for highly variable hippocampal anatomy. In addition, we plan to apply HippMetric to human development and disease progression studies to characterize normative trajectories and subfield-specific alterations in disorders such as AD. These efforts may facilitate the development of new imaging biomarkers, ultimately supporting early diagnosis, risk stratification, and therapeutic development.

## Data Availability Statement

Data supporting the findings of this study were collected from Alzheimer's Disease Neuroimaging Initiative (ADNI) and Greater-Bay-Area Healthy Aging Brain Study (GHBAS). The ADNI database is publicly available at www.loni.ucla.edu/ADNI/. The raw data of GHABS cohorts are not currently publicly available due to the requirement of local government policy on the export and sharing of clinical and population data. However, the data used in this study are available from T.G. (tengfei.guo@szbl.ac.cn) upon request by any qualified academic investigator, subject to a data use agreement for the sole purpose of replicating procedures and results. The inclusion and exclusion criteria were listed in the "Datasets" section of the supplementary materials. All code for this project, including the template, is available at https://github.com/calliegao/HippMetric-case-by-case.

## Code Availability Statement

The HippMetric toolbox is freely available at: https://github.com/calliegao/HippMetric-case-by-case. The toolbox provides a modular Python implementation for hippocampal morphometric modeling and quantification along its longitudinal lamellar axis. The package includes end-to-end scripts for hippocampal coordinate construction, cross-subject and longitudinal alignment, as well as shape-based measurements of thickness, width, and curvature.

In addition to the core codebase, the repository contains example datasets, intermediate outputs, and pre-configured parameter files to facilitate the replication of the pipeline evaluations presented in this work. Comprehensive documentation and usage tutorials are available at: https://HippMetric-case-by-case.readthedocs.io.

## Acknowledgement

This study was funded by the National Natural Science Foundation of China (82422027, and U24A20340), National Key Research and Development Program of China (2023YFC3605400), Guangdong Basic and Applied Basic Science Foundation for Distinguished Young Scholars (2023B1515020113), Shenzhen Bay Laboratory (S241101004-1). We thank all the GHABS, ADNI participants, and staff for their immense contributions to data collection. The Shenzhen Bay Laboratory supercomputing center supported the imaging processing. The data collection and sharing for ADNI was funded by the Alzheimer's Disease Neuroimaging Initiative (ADNI) (National Institutes of Health Grant U01 AG024904) and DOD ADNI (Department of Defense award number W81XWH-12-2-0012). ADNI is funded by the National Institute on Aging and the National Institute of Biomedical Imaging and Bioengineering and through generous contributions from the following: AbbVie; Alzheimer's Association; Alzheimer's Drug Discovery Foundation; Araclon Biotech; BioClinica, Inc.; Biogen; Bristol-Myers Squibb Company; CereSpir, Inc.; Eisai Inc.;

Elan Pharmaceuticals, Inc.; Eli Lilly and Company; EuroImmun; F. Hoffmann-La Roche Ltd and its affiliated company Genentech, Inc.; Fujirebio; GE Healthcare; IXICO Ltd.; Janssen Alzheimer Immunotherapy Research & Development, LLC; Johnson & Johnson Pharmaceutical Research & Development LLC; Lumosity; Lundbeck; Merck & Co., Inc.; Meso Scale Diagnostics, LLC; NeuroRx Research; Neurotrack Technologies; Novartis Pharmaceuticals Corporation; Pfizer Inc.; Piramal Imaging; Servier; Takeda Pharmaceutical Company; and Transition Therapeutics. The Canadian Institutes of Health Research is providing funds to support ADNI clinical sites in Canada. Private sector contributions are facilitated by the Foundation for the National Institutes of Health (www.fnih.org). The grantee organization is the Northern California Institute for Research and Education, and the study is coordinated by the Alzheimer's Disease Cooperative Study at the University of California, San Diego. ADNI data are disseminated by the Laboratory for Neuro Imaging at the University of Southern California.

## CRediT authorship contribution statement

Na Gao: Writing - original draft, Project administration, Methodology, Formal analysis, Visualization, Conceptualization. XIngyu Hao: Investigation, Methodology, Formal analysis, Visualization. Yanwu Yang: Writing - review & editing. Zhiyuan Liu: Writing - review & editing. Anqi Li: Investigation, Formal analysis. Zhengbo He: Investigation, Formal analysis. Li Liang: Investigation, Formal analysis. Chenfei Ye: Writing - review & editing. Ting Ma: Writing - review & editing, Conceptualization. Tengfei Guo: Writing - review & editing, Supervision, Funding acquisition.

## Competing interests

The authors declare no competing interests.